\documentclass[journal]{IEEEtran}

\usepackage{amsmath,amssymb,amsfonts}
\usepackage{graphicx}
\usepackage{listings}
\usepackage{microtype}
\usepackage{colortbl}
\usepackage{color}
\usepackage[tight,footnotesize]{subfigure}
\usepackage{todonotes}
\usepackage{url}
\usepackage{textcomp}
\usepackage{hyperref}
\usepackage{tcolorbox}
\usepackage{multirow}

\lstdefinelanguage{m3}{
	basicstyle=\ttfamily\scriptsize,
	keywordstyle=\bfseries,
	keywords={m3,declarations,methodInvocation},
	literate={<-}{$\leftarrow$}{1},
	tabsize=2,
	alsoletter={-}
}

\usepackage{xspace}
\newcommand*{\ie}{i.e.,\@\xspace}
\newcommand*{\eg}{e.g.,\@\xspace}

\makeatletter
\newcommand*{\etc}{%
	\@ifnextchar{.}%
	{etc}%
	{etc.\@\xspace}%
}
\makeatother
\newcommand*{\etal}{\emph{et~al}.\@\xspace}

\definecolor{verylightgray}{gray}{0.98}

\newcommand{\rqfirst}{\emph{\textbf{RQ$_1$:} Is the approach able to reconstruct bus routes from GPS traces?}~}% Are we able to retrieve the bus route from the GPS traces compared to those of Hierarchical Clustering  of the classification
\newcommand{\rqsecond}{\emph{\textbf{RQ$_2$:} Can the approach be used to predict bus stops?}~} %deduce missing information
\newcommand{\rqthird}{\emph{\textbf{RQ$_3$}}: \emph{Is the approach efficient in terms of timing?}~} %\EP able to provide consistent recommendations in reasonable time

\definecolor{lightgray}{gray}{0.85}
\definecolor{verylightgray}{gray}{0.95}
\definecolor{darkblue}{rgb}{0.1,0.6,1.0}
\definecolor{red}{rgb}{1.0, 0.01, 0.24}
\definecolor{gray}{rgb}{0.8, 0.8, 0.8}

\begin{document}

%====================== We should discuss to select a suitable title for our paper======================
%======================%======================%======================

%\title{Predicting Transit Feed Specification with LSTM Recurrent Neural Networks}

%\title{Making Unavailable Transit Feed Specification Available with Recurrent Neural Networks}

%\title{Making Available Unavailable Transit Feed Specification with Recurrent Neural Networks}

%\title{Unavailable Transit Feed Specification: Making it Available with Recurrent Neural Networks}

\title{Unavailable Transit Feed Specification: Making It Available with Recurrent Neural Networks}

\author{Ludovico~Iovino,
	Phuong~T.~Nguyen,
	Amleto~Di~Salle,
	Francesco~Gallo,
	and~Michele~Flammini% <-this % stops a space
	\thanks{L. Iovino and M. Flammini are with the Gran Sasso Science Institute, L'Aquila,
		    67100, IT e-mail: \{ludovico.iovino, michele.flammini\}@gssi.it}
	\thanks{P. T. Nguyen, A. Di~Salle, and F. Gallo are with the Department
		of Information Engineering, Computer Science and Mathematics, University of L'Aquila, L'Aquila, 67100 IT e-mail: \{phuong.nguyen, amleto.disalle, francesco.gallo\}@univaq.it}% <-this % stops a space
	%\thanks{Manuscript received MMMM DD, 2020; revised MMMM DD, 2020.}
	}

%\author{Michael~Shell,~\IEEEmembership{Member,~IEEE,}
%	John~Doe,~\IEEEmembership{Fellow,~OSA,}
%	and~Jane~Doe,~\IEEEmembership{Life~Fellow,~IEEE}% <-this % stops a space
%	\thanks{M. Shell was with the Department
%		of Electrical and Computer Engineering, Georgia Institute of Technology, Atlanta,
%		GA, 30332 USA e-mail: (see http://www.michaelshell.org/contact.html).}% <-this % stops a space
%	\thanks{J. Doe and J. Doe are with Anonymous University.}% <-this % stops a space
%	\thanks{Manuscript received April 19, 2005; revised August 26, 2015.}}

%\author{\IEEEauthorblockN{Ludovico Iovino}
%\IEEEauthorblockA{Gran Sasso Science Institute, Italy \\
%Email: ludovico.iovino@gssi.it}
%\and
%\IEEEauthorblockN{Phuong T. Nguyen, Francesco Gallo, Amleto Di Salle}
%\IEEEauthorblockA{Universit\`a degli Studi dell'Aquila, Italy \\
%Email: \{phuong.nguyen, francesco.gallo, amleto.disalle\}@univaq.it}}


\markboth{Journal of IEEE Transactions on Intelligent Transportation Systems,~Vol.~xx, No.~x, XXXX~2021}%
{Shell \MakeLowercase{\textit{et al.}}: Making the Unavailable Available in Transit Feed Specification}

\maketitle

\begin{abstract}

%In the European Union, nearly 60 billion travels are performed through public transport. At the same time, only 19\% of all Europeans claim they use public transport regularly. 

%The mode of public transportation most frequently used in Europe highlights that the European inhabitants use buses in ca. 56\% of all public transport travels. 
%In Europe, traveling with buses accounts for ca. 56\% of all public transport means. 
Studies on public transportation in Europe suggest that European inhabitants use buses in ca. 56\% of all public transport travels. One of the critical factors affecting such a percentage and more, in general, the demand for public transport services, with an 
increasing reluctance to use them, is their quality. End-users can perceive quality from various perspectives, including the availability of information, \ie the access to details about the transit and the provided services. The approach proposed in this paper, using innovative methodologies resorting on data mining and machine learning techniques, aims to make available the unavailable data about public transport. In particular, by mining GPS traces, we manage to reconstruct the complete transit graph of public transport. The approach has been successfully validated on a real dataset collected from the local bus system of the city of L'Aquila (Italy). The experimental results demonstrate that the proposed approach and implemented framework are both effective and efficient, thus being ready for deployment. %can be used in the field
%showing the effectiveness of our solution.

% testing the proposed approach with the data of

\end{abstract}


% \fi
\IEEEpeerreviewmaketitle

\section{Introduction}\label{sec:introduction}

%<<<<<<< HEAD
%In the European Union, nearly 60 billion travels are performed by means of public transport\PN{@Ludovico: Annually, monthly (??). We also need a reference for this claim}. At the same time, only 19\% of all Europeans claim they use public transport on a regular basis. The mode of public transport most frequently used in Europe highlights that buses are used by the European inhabitants in ca. 56\% of all public transport travels~\cite{qtraspo}. One of the key factors affecting the demand for public transport services, which increases the reluctance in using them, is given by their quality. This quality can be perceived by end-users from various points of view, including the availability of information, \ie the access to information about transit and the provided services. Among other advantages, public transport is an effective means for reducing the number of journeys made by private cars, and thus cutting down traffic jams as well as air pollution. Facilitating public transport (\emph{to be continued}).
%=======
Among other advantages, public transport is an effective means for reducing the number of journeys made by private cars and thus helping to cut down traffic jams as well as air pollution. Facilitating public transport has a profound impact on both the economy and the living environment. In the European Union, around 60 billion trips are made annually with public transport.\footnote{Quality analysis of the public transport systems, available at \url{http://interconnect.one}} A recent report reveals that buses in Europe are used in ca. 56\% of all public transport travels~\cite{qtraspo}. At the same time, only 19\% of all Europeans claim they use public transport regularly. Among the others, the quality of service is a crucial factor affecting the demand and willingness to use them. Such a quality can be perceived by end-users from various perspectives, including the availability of information, \eg access to details about the transit and the provided services.

% The mode of public transport most frequently used in Europe highlights that 
%\PN{to be continued.}
%>>>>>>> 0d4805ba8e1b11199058ffdd6edd266b6c6eea4e

%One of the key factors 
%economic ad

%And it didn’t even take much longer, since most cities coordinate their train and bus systems pretty efficiently.
%, and facilitating public.
%It is important to remark that enhancing and improving public transport means reducing the number of travels by private cars, thus contributing to the reduction of the congestion, noise and air pollution costs as well as the number of road accidents. 
%Public Transport is a good way to reduce congestion and environment and health-harming emissions in urban areas, especially when they run on alternative, cleaner fuels. The European Commission strongly encourages the use of public transport as part of the mix of modes which each person living or working in a city can use. For example, parking one's car outside of the city centre at a railway or bus station. https://ec.europa.eu/transport/themes/urban/urban_mobility/urban_mobility_actions/public_transport_en

%Buses offer fewer cues for stops, so it’s even more important to stay alert. Have a sense of how long the trip is going to take. As you ride, follow along the route on your map, looking for landmarks along the way: monuments, bridges, major cross streets, and so on. Some buses pull over at every stop, while others only stop by request. If in doubt, look for a pull cord or a button with the local word for “stop.”

When it comes to Smart Cities, \emph{infomobility} represents one of the key required services, and the development of corresponding solutions is part of a continuous process employed to handle environmental and security problems. The challenge is to offer tourists and local citizens a simple and intuitive service, allowing them to seamlessly obtain information related to public transportation %and related aspects,
without any complex computer skills. Thus, among the various crucial aspects in the process of developing a smart city, infomobility has garnered considerable attention both from academia and industry.
% is certainly one of the most cited, and it deserves attention, being connected to all the other aspects.
%what time of data 
%First of all, transit agencies must offer information systems and this can be seen as a technology transfer process, especially in Italy, where many systems for providing transport information are not at the current standard level. 
Public transit agencies, while having as primary goal of improving on-time performance and offering a better service, should also provide users with suitable facilities, like trip planners, i.e., smart travel assistance tools which give pre-trip information for origin-destination pairs~\cite{BOROLE2013775}. A travel planner system gathers information such as the road network and the buses’ schedule from one or many transport agencies. It is available to commuters usually through a map-based application. %Moreover, transit services must have a reliable service in terms of frequency and travel time to attract more passengers. %For this reason, 
Real-time transit information is a cost-effective feature that can improve the perception of reliability from the users' perspective~\cite{WATKINS2011839}. It includes real-time details of bus itineraries or waiting time. This type of information can be used as input for other services, \eg next bus countdown dashboard, notifications of arrival, notification of interruption or traffic jam. Eventually, these facilities can be provided to end-users via website, emails, text messaging, or mobile apps. Such services are quite appealing for passengers and increase their level of satisfaction, as they help them to figure out the estimated time to reach their destination, or the position and waiting time for a specific bus, or other contextual situations of the public transport.
%and can invest the remaining time in private matters. 
%For this reason, 
These features are fundamental to increasing the level of satisfaction of passengers. 

% can drastically affect user perception of
%open problem

The unavailability of data is the first obstacle to a smart infomobility system. This situation is still a reality for a lot of small cities, where different companies manage public transport, often with somewhat obsolete software procedures. Data incompatibility and erosion are the leading cause of this lack of data, and infomobility systems are not well supported in this matter.
This gives rise to the following problems: %If data, \eg road network and buses schedule is not available or result incompatible the result is that: 
\emph{(i)} infomobility systems, including travel planners cannot be available, or if available they are not in line with the updates of the bus service provided by the transport agency; \emph{(ii)} real-time tracking is not helpful since the system cannot link the tracked bus with the schedule and travel which users are looking for. Recently, there have been a number of available open-source projects to provide passenger information and transportation network analysis services. An example is OpenTripPlanner\footnote{\url{https://www.opentripplanner.org}} (OTP), a system that offers features to find multi-modal itineraries combining transit, pedestrian, bicycle, and car segments through networks built from the widely available open standard Open Street Map (OSM) and Google Transit Feed Specification (GTFS) data. The project offers a ready-to-use solution to transport agencies that want to provide infomobility services to customers, without investing money for developing brand new applications. Nevertheless, since data is not available or is in an incompatible format, the system cannot work. 

%Machine Learning is a data analysis methodology that automates analytical model building. 
%It is based on the concept that systems can learn from data, identify patterns and make decisions using existing dataset. 

In recent years, we have witnessed a rise in the application of Machine Learning (ML) algorithms in different domains. 
%The recent development of several disruptive Machine Learning (ML) algorithms holds promise for success in different application domains. 
%Machine Learning 
Such algorithms are capable of simulating humans' learning activities, aiming to acquire real-world knowledge autonomously~\cite{PORTUGAL2018205} by generalizing from concrete examples%. In other words, they are able to identify patterns and make decisions by means of data, without being manually coded or intervened by humans
~\cite{Domingos:2012:FUT:2347736.2347755},\cite{doi:10.1080/21693277.2016.1192517}. %Thanks to this characteristic, they have been widely applied in several application domains, such as Web search%by learning from a user’s long-term search history
%~\cite{Sontag:2012:PMP:2124295.2124348}, recommender systems %, ML algorithms demonstrate their superiority by analyzing sentiment with ensemble techniques in social applications
%~\cite{ARAQUE2017236}, %health, %or allowing systems to learn from various profiles, thus boosting up the recommendation outcomes
%\cite{PORTUGAL2018205}. Machine Learning algorithms are also indispensable to the controlling of 
%self-driving cars~\cite{doi:10.1177/0306312717741687}, to name a few.
In this work, aiming at predicting transit feed specifications through available training data, we present an ML-based approach to enable GTFS-based travel planners. %given that no data is available or usable. 
We designed and implemented a framework that processes real-time position data transmitted from buses to predict routes and trips, thus providing practical information related to bus itineraries. 
%More importantly, we also succeeded in retrieving bus stops and fixing wrongly represented trips files provided by the involved travel agency. %as a baseline. 
A reverse engineering process has then been used to obtain the OTP graph and enable all the above-mentioned features, letting them available in our infomobility system. Such a process is generic and exempt from manual encoding, it can be easily replicated in different urban scenarios and can be integrated to nowadays trip planning systems. We tested our approach using a real-world dataset collected from 10 buses equipped with GPS devices. To the best of our knowledge, our work is the first one that can effectively and efficiently provide two different types of recommendations at the same time, namely \emph{(i)} prediction of bus trajectories; and \emph{(ii)} prediction of bus stops.
Thus, the main contributions of our work are summarized as follows: \emph{(i)} a prototype built on top of an long short-term memory recurrent neural network (LSTM) to predict and retrieve bus trajectories and bus stops; \emph{(ii)} an empirical study to evaluate the approach on a real dataset provided by the agency managing the local transportation system of the city of L'Aquila, Italy; \emph{(iii)} a light and robust framework, resulting in an application that can run on mobile devices to provide infomobility information, even in an offline fashion, without having to resort to an Internet connection (see the final discussions for more details).

% in order to understand which schedule is serving %using GPS-traces 
\vspace{.1cm}
\noindent
\textbf{Structure of the paper.} The paper is organized as follows.
Section~\ref{sec:related} reviews some related studies and associates them with our work, while Section~\ref{sec:MotivatingExample} provides a motivating example related to the exposed problem. Section~\ref{sec:background} presents a background to our work, together with an introduction to recurrent neural networks and the long short-term memory technique, as well as to the Google Transit Feed Specification. We present our proposed approach in Section~\ref{sec:approach}. %Section~\ref{sec:background} introduces some basic concepts we used in our approach.
Section~\ref{sec:evaluation} explains in detail the material and methods used to evaluate the approach. Section~\ref{sec:results} analyzes the %experimental 
results obtained from the evaluation. Finally, Section~\ref{sec:conclusione} sketches future work and concludes the paper.

\section{Related Work}\label{sec:related}
%FROM GALLO: The use of emerging technologies for data collection, such as location-aware smartphones, have become increasingly popular. Much work in this field has concentrated on methods of inferring characteristics of the trip from passively or dynamically collected data. Travel demand forecasting and modelling are key tools in evaluating transport plans, initiatives, and policies in urban areas in this way. This section offers a non-exhaustive overview of machine learning-based approaches.

In this work, we present a solution to the prediction of bus trajectories exploiting an LSTM. In particular, the system infers the next positions and stops of a bus while travelling, based on its recent locations. Our work is mainly related to approaches and methods developed for deducing characteristics of trips from passively or dynamically collected data. This section reviews notable studies in the same domain and relates them to our work.

%To this end, we propose a non-exhaustive overview of machine learning-based approaches used into the info mobility domain.
\noindent
Several approaches have been proposed to predict, plan and classify trip information through GPS data using ML techniques~\cite{doi:10.1080/03081060.2017.1314502}, ~\cite{Ermagun:2017}, ~\cite{Lin2017DeepGM}, ~\cite{Zhang0SZ18}. In particular, Dalumpines \etal~\cite{doi:10.1080/03081060.2017.1314502} 
present a four-stage method to automatically extract activity episodes from GPS data (latitude, longitude, and time). Statistical descriptors are derived from the extracted episodes, and then employed within a multinomial logit (MNL) model to classify extracted episodes into different types: stop, car, walk, bus, and other (travel) episodes. Ermagun \etal~\cite{Ermagun:2017} develop prediction models to forecast trip purposes, \eg education, eat out, mainly based on online location-based search and discovery services, \eg Google Places API. Probabilistic and machine learning methods have been exploited, \eg nested logic and random forest, to predict trip purposes in real-time upon the completion of the trip. Lin \etal~\cite{Lin2017DeepGM} propose a two-step generative modeling framework to carry out the most common transportation planning tasks. In the first step, an Input-Output Hidden Markov Model (IO-HMM) is used to sequence and label activities from the cellular data network. %(call detail records). 
Then, these activities are used to train an LSTM network which generates synthetic travel plans. Proposed by Zhang \etal~\cite{Zhang0SZ18}, DeepTravel is composed of two layers, the feature representation layer and the prediction layer. The first layer uses different strategies to extract different features, \ie spatial and temporal embedding, driving state, short-term and long-term traffic, from GPS data related to a travel path. Then, the second layer estimates the travel time of a query path through a bidirectional LSTM and a dual interval loss mechanism for auxiliary supervision.

There exist some approaches and models to prediction of bus arrival times. The most recent techniques use RNNs to capture long-range dependencies. For example, an RNN with LSTM block~\cite{8516374} has been used to improve the forecast for a station by taking into account the correlated multiple passed stations. Agafonov and Yumaganov~\cite{Agafonov2019222} adopt RNN with LSTM architecture and heterogeneous information, such as real-time and statistical data concerning vehicle traffic, to predict public transport arrival time. Liu \etal~\cite{Liu202011917} propose a hybrid neural network by using two prediction models, an LSTM model and an artificial neural network. The network solves the bus-station arrival prediction problem by employing spatial-temporal feature vectors. Zhou \etal~\cite{Zhou2019} adopt an RNN that employs a set of dynamic factors, \eg passengers number, dwell time, bus driving efficiency, to predict bus arrival times. Moreover, the authors introduce an attention mechanism to select the most relevant factors from heterogeneous information adaptively. In contrast, our approach can predict bus routes, trips and bus stops, where trips are individual components of routes. 

Soares \etal~\cite{9013316} propose an approach to classification of travel modes and accommodation of transportation infrastructures and services to meet citizens' needs. A preprocessing phase was conducted to extract the required features, which were then fed as input for an LSTM recurrent neural network. The approach was evaluated on a dataset collected from volunteers in Bologna, Italy. Our proposed model is also built on top of an LSTM, however, it is used to predict both bus routes and bus stops.
Yazdizadeh \etal~\cite{YAZDIZADEH201982} propose a framework exploiting machine learning techniques to recognize complete trip information based on smartphone location data and combining them with online GTFS data and Foursquare\footnote{\url{http://www.foursquare.com}} data. In particular, the approach uses data to train and validate three random forest models, which allow to predict mode of transport, transit itinerary and user activities. Differently from our approach, they focus on deducing trip characteristics, rather than predicting bus routes, trips and bus stops by means of LSTM. 
In some recent studies, finite-time zeroing neural networks (ZNN) have been conceptualized to find the accurate solution of Lyapunov equation, given that no noises exist~\cite{8648298}. Attempts have been made to enhance the convergence speed of ZNN and mitigate noises in real applications. Similarly, a new ZNN using a versatile activation function (VAF) has been proposed to solve time-dependent matrix inversion~\cite{XIAO2019124}. We anticipate that the application of such techniques can help improve the prediction accuracy in the context of predicting bus trajectories. We consider this our future work.

%. To the best of our knowledge, our work is the first one that can
%the proposed ZNN model not only converges to zero within a predefined finite time but also tolerates several noises in solving the time-dependent matrix inversion, and thus called new noise-tolerant ZNN (NNTZNN) model ~\cite{XIAO2019124}
 
%, in this paper, two robust nonlinear zeroing neural networks (RNZNNs) are designed by adding two novel nonlinear activation functions (AFs) for finding the solution of the Lyapunov equation in the presence of various noises~\cite{8648298}.

%In our future work, we will consider the error reduction aspects following the 
%A model~\cite{XIAO2019124}. 

%AMLETO: REMOVED SINCE IT IS NOT RELATED
%In a recent work~\cite{Assemi202089}, the authors propose and evaluate the application of a probabilistic approach using neural networks to reduce the errors during inferring of the alighting stop in public transport origin-destination estimation, based on smart card data. Compared to our approach, they use data from the user's smart-card, which must be complemented by information about the environment surrounding the alighting stop and physical characteristics of the user, such as leg length.

%START GALLO
Jabamony \etal~\cite{Jabamony2020312} propose an approach to improve the performance of the intelligent public transport with IoT enabled system~\cite{Li2019} to give a transport travel time prediction at any distance along with the route and the arrival time of the bus to the particular bus stops. The approach is based on an artificial neural network which has been trained with different traffic parameters and environmental conditions. Meanwhile our approach is built on top of an LSTM and it makes use of only GPS data.
Petersen \etal~\cite{Petersen2019426} use a simple standard output, that most automatic vehicle location systems use in the public transport industry, as input to a multi-output, multi-time-step, deep neural network combining CNN and LSTM for bus travel prediction.

%END GALLO

%============================\textbf{Comment 1.1:}============================
By investigating related studies, we see that while they address a problem similar to ours, as well as they solve different technical aspects, there is still some room for improvement. First, there are issues with either efficiency or effectiveness, \eg some tools can provide accurate predictions, but they still suffer from a high computational complexity. Second, the prediction of bus trips and bus stops has generally been studied in separate work. In this way, we propose a practical solution %to the problem. First, 
%our conceived framework is effective as it can provide 
by providing forecasts of both routes (trips) and stops. More importantly, the framework is also efficient in terms of timing, \ie it can provide real-time recommendations while the bus is en route. In this respect, we assume that our proposed solution fills the gap by introducing some significant deltas.

\section{Motivating Example} \label{sec:MotivatingExample}

%===================the following paragraph needs to be completely improved.
%underdeveloped
% substantial

%Despite recent technological advancements, 
Despite recent breakthroughs in information and communications technology, 
many transport agencies still provide services to passengers in an old-fashion manner. %This is the case %Technological advancements are done only in the direction of the transport agency and not to provide new services to the user. 
%for many Italian cities, where there are multiple examples as highlighted in the ranking of smart cities published recently~\cite{EY-smartcities}. %for %for %smart cities in Italy 
According to a recent report~\cite{EY-smartcities}, %While there are cities on top, 
many Italian cities are not ready for the Smart Cities concept, since the information systems to manage the daily public transportation are not yet on a par with the basic requirements. Given their circumstances, there is a pressing need to adapt and enhance the existing infrastructures.%reshuffle. %, they have %of them are located on the bottom of the ranking, indicating a a transportation %infrastructure of a low standard, which needs a thorough overhaul. %Most of the cases this is due to the user perception of the provided services, often wholly absent ...

As a typical example, let us describe the situation of the city of L'Aquila till a few years ago. L'Aquila is a city in central Italy, located in a mountainous area, around 700 meters above the sea level, and it has suffered frequent earthquakes of different magnitudes in its history, dating back to thousands of years. Partly due to these geographical characteristics, the city's public transportation system relies only on buses. The city was a typical example of a working public transport system but with a limited infomobility service. The transport agency, named AMA\footnote{\url{http://www.ama.laquila.it}}, provided the timetables and related information as PDF documents through its website, even if it is now undergoing a process of migrating to new technologies. Such a process needs still to be completed. As a matter of fact, the main national opendata portal is not yet in line with its transportation service.\footnote{In the \href{http://www.datiopen.it/it/opendata/Mappa_fermate_autobus_in_Italia}{Italian Open Data portal}, there is limited data available for L'Aquila.} 

\begin{figure}[h!]
	\begin{center}
		\vspace{-.2cm}
		\includegraphics[width=0.80\columnwidth]{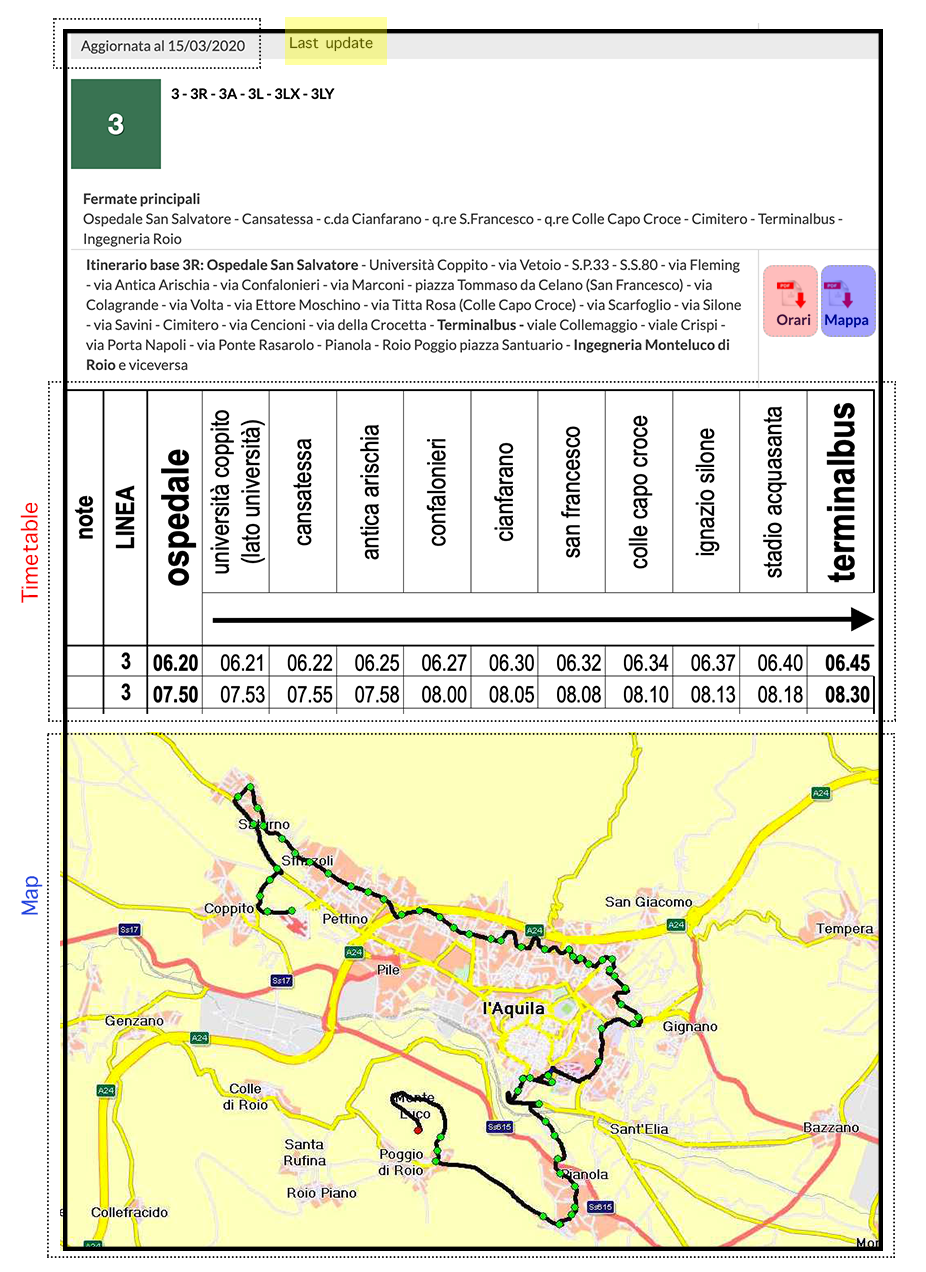}		
		\caption{Timetable of Line 3.} %for public transport in L'Aquila with focus on 
		\vspace{-.4cm}
		\label{fig:timetable}
	\end{center}
\end{figure}

%An example of data made available is reported in 
An example of the available information in the above mentioned PDF files is shown in Fig.~\ref{fig:timetable}, which provides details of Line 3 of the local transportation service, obtained combining different sources. In particular, the top left corner of the figure specifies the last update time, followed by
%The top of the screenshot features %of the website of the transit agency reports 
the bus stops of the selected route, %as well as the last update time. 
detailed in the timeline in the center. Eventually, at the bottom there is a picture plotting the map corresponding to the route. %of the covered path is represented. 
The timeline is linked with a PDF document, while the map is linked with a JPG/PNG picture. The text \emph{``Last update''} redirects commuters to the interim information, which also implies that the document %has to be updated for every change the agency applies. This creates the impression that the document 
may not be up to date with the current transport timetable, with all the consequent issues. For example, it can happen that a bus does not come at a specified time, due to recent changes of the timetable.

If the software involved in the management of the schedules used by the local agency does not provide a front-end platform for travel planning, nor it supports the export in a format that can be processed by open-source software, managers are stuck in a situation of vendor lock-in or they are forced to a technology switch, with all the consequent additional costs. 

A probable solution could be processing documents and converting data to the Google Transit Feed Specification (GTFS)\footnote{\url{https://developers.google.com/transit/gtfs}}, that is a standard format for public transportation schedules and the associated geographical information (see Section~\ref{sec:gtfs} for further details). GTFS allows public transit agencies to publish their transit data as well as developers to write applications that consume such data in an interoperable manner. %Multiple projects provide 
For instance, various tools can be used to convert Excel spreadsheets into GTFS feed. %but as one can imagine, 
Nevertheless, such as an input format is not universal, and the transformation process needs to be customized continuously. This operation is time-consuming, error-prone, and needs to be replicated every time the schedule is updated. 
%\begin{figure*}[h!]
%	\begin{center}
%		\includegraphics[width=2\columnwidth]{figs/OTP.png}
%		\caption{Screenshot of OTP with imported data of Line 3R.}
%		\label{fig:OTP}
%	\end{center}
%\end{figure*}

For this reason, we abandoned the idea of building a communication layer from the website and data provided as in Fig.~\ref{fig:timetable} to GTFS, that can be used to directly feed one of the %open source 
platforms, \eg OTP or Google. This is also particularly overcomplicated when multiple transit agencies are involved, and the management is delegated to different %multiple supporting 
companies. Indeed, this process is not generic and needs to be re-executed for all data sources and export format. Moreover, in the current local situation, HTML scraping is the only way to grab this content, given that the data owner does not provide %have the resources to expose it as 
a machine-readable service, \eg RESTful Web services.

To generate GTFS feeds, one may think of a static program that analyzes coordinates by connecting to a database populated with GPS data. However, such a software component requires time, effort and costs to be developed and might be computationally expensive when dealing with several bus lines. Moreover, the static program may fail, considering the fact that GPS devices do not always report coordinates which exactly match with those %that have been
stored in the database. Finally, a static solution would have difficulties in adjusting to the different traffic and driver behaviour situations. In this sense, we need a practical solution to transcend the limitations.

%To show a possible outcome of this operation, %what would be the result of the import for this data, 
%we imported a portion of the published Excel file and converted them into the GTFS format. The result of this manual operation is depicted in Fig.~\ref{fig:OTP}, where the same bus line reported in the JPG file shown in Fig.~\ref{fig:timetable}. Importing GTFS data provided within the OTP platform allows a commuter to plan a trip, for instance within the reported line: by specifying the required source and destination, the passenger can query the dedicated platform which returns the corresponding path together with its bus stops on the map.

%=================the need for technical support=================
%in Section~\ref{sec:approach} 

%Given the circumstances, 
To automatize the process, we came up with the idea of exploiting a well-founded ML technique to provide commuters %in the city of L'Aquila 
with practical information related to trips and routes. In particular, in our solution we train a long short-term memory recurrent neural network by means of %For all these scenarios where data is not available, we propose an automatic approach that works on top of machine learning algorithms by means of 
real-time data collected from GPS trackers, and process it %mounted on buses 
to deduce missing information and eventually to export proper GTFS transit feeds. %  feeds.
%This process will enable the usage of platforms for exposing info to the users, for instance OTP Platform. 
Once being properly realized, such a process can be used to enable open trip planners with all the essential functionalities, in every urban scenario, regardless of their information systems and technologies to manage public transportation.

%In the next subsection, we review the main technical background to pave the way for further presentations.

%behind the ...  We review some technical background of 

%Section~\ref{sec:approach} highlights the proposed approach.
%is demonstrated in section~\ref{sec:approach}, whereas the background needed to understand the section is shown in section~\ref{sec:background}.

\section{Background}\label{sec:background}
In this section, we briefly introduce the GTFS format and recurrent neural networks, together with the related long short-term memory technique. %Moreover, we also 

\vspace{-.3cm}
\subsection{Google Transit Feed Specification}\label{sec:gtfs}
Data availability is among the essential requirements to enable the use of multi-modal trip planning systems. To this end, the Google Transit Feed Specification - GTFS has been proposed as a common format for exchanging information among platforms.
%For this reason multiple platforms went to a common direction of having an interchange format as GTFS. 
%OLD VERSION: The General Transit Feed Specification (GTFS) is a data specification used by public transit agencies to publish their transit data in a format that can be processed by other software specifications. Novadays, the GTFS format is used by thousands of public transport providers\footnote{https://gtfs.org/}. A GTFS feed is composed of textual files compressed in ZIP files, and each file models a particular aspect of transit information: stops, routes, trips, and additional schedule data. A Web server offering GTFS allows third parties to connect and make use of the available services.
It is a data standard used by public transit agencies to publish their transit data in a format that can be processed by other software systems. Nowadays, it is widely used by several public transport providers around the world.
A GTFS feed consists of textual files (in CSV format) compressed in a ZIP file, each file modeling a particular aspect of the transit information: agency, stops, routes, trips, and additional schedule data. For illustration purposes, in Table~\ref{tab:gtfs-feed-example} we display an excerpt of the GTFS feed specification related to the motivating example of Section~\ref{sec:MotivatingExample}. In particular, the \textit{agency.txt} file contains data related to the agency, \textit{stops.txt} reports the latitude and longitude of each stop, \textit{routes.txt} includes the transit routes, and \textit{trips.txt} provides trips for each route (a sequence of two or more stops related to a specific period forms a trip).
\begin{table}[!ht]
	\small
	\centering		
	\footnotesize
	\scriptsize
	\vspace{-.4cm}
	\caption{An excerpt of the GTFS feed for the motivating example.} % used in the paper
	\begin{tabular}{|p{0.4cm}|p{2.7cm}|p{2.7cm}|p{1.1cm}|}	\hline
		\rowcolor{verylightgray} 
		\multicolumn{4}{|c|}{\textbf{agency.txt}}\\ \hline  
		\textbf{id} & \textbf{name} & \textbf{url} & \textbf{timezone} \\ \hline 
		AMA& Azienda Mobilit\`{a} L'Aquila & \url{http://www.ama.laquila.it/} & CEST  \\ 
		\hline
	\end{tabular}		
	\begin{tabular}{|p{0.4cm}|p{2.1cm}|p{1.5cm}|p{1.1cm}|p{0.3cm}|p{0.6cm}|}	\hline
		\rowcolor{verylightgray} 
		\multicolumn{6}{|c|}{\textbf{stops.txt}}\\ \hline  
		\textbf{id} & \textbf{name} & \textbf{lat} & \textbf{lon} & \textbf{type} & \textbf{parent} \\ \hline 
		H1&Universit\`{a} Coppito&42.367679&13.352023&2&- \\ \hline
		S2&S.P. 33 - S.S. 80&42.369472&13.346469&2&H1 \\ \hline
		S3&Via Fleming&42.383699&13.342006&2&H1 \\ \hline
		S4&Via Antica Arischia&42.380922&13.346931&2&H1 \\ \hline
		S4&Via Antica Arischia&42.373691&13.358209&2&H1 \\ \hline		
	\end{tabular}		
	\begin{tabular}{|p{0.4cm}|p{1.3cm}|p{1.7cm}|p{2.6cm}|p{0.4cm}|}	\hline
		\rowcolor{verylightgray} 
		\multicolumn{5}{|c|}{\textbf{routes.txt}}\\ \hline  
		\textbf{id} & \textbf{agency id} & \textbf{short name} & \textbf{long name}& \textbf{type} \\ \hline 
		S3&AMA&3&Ospedale&3 \\ \hline 
	\end{tabular}		
	\begin{tabular}{|p{1.1cm}|p{1.5cm}|p{1.9cm}|p{0.9cm}|p{2.7cm}|p{1.0cm}|}	\hline
		\rowcolor{verylightgray} 
		\multicolumn{6}{|c|}{\textbf{trips.txt}}\\ \hline  
		\textbf{route id} & \textbf{service id} & \textbf{trip ip} & \multicolumn{2}{c|}{\textbf{headsign}}& \textbf{block id} \\ \hline 
		L3&Feriali&L3FERIALI1&\multicolumn{2}{c|}{Ospedale}&1\\ \hline 	
	\end{tabular}
	\label{tab:gtfs-feed-example}
	\vspace{-.2cm}
\end{table}

%It is worth noting that t
The data in Table~\ref{tab:gtfs-feed-example} is an excerpt of what we planned to obtain with the proposed approach, which will be detailed in Section~\ref{sec:approach}.

\vspace{-.3cm}
\subsection{Recurrent Neural Networks and LSTMs} \label{sec:RNNandLSTM}

%\input{src/LSTM}

%==========================================The whole paragraph needs to be paraphrased==========================================

%The ongoing adoption of 
%The proliferation of Machine Learning (ML) techniques in recent years has enabled a numerous number of applications in various domains. 

%====================================The following paragraph should be move to the Related Work section====================================
%Recently, a number of studies have been dedicated to apply ML techniques to solve different issues in the agriculture sector~\cite{KAMILARIS201870}. For fruit classification, various approaches have been proposed by applying deep learning techniques~\cite{KAMILARIS201870,REHMAN2019585,MUANGPRATHUB2019467}. Still, many existing approaches require long training/testing time or suffer a limited accuracy, which may impede real-time usage. Given the circumstances, it is necessary to study and find an appropriate model for fruit classification.
%==========================================================================================================================================
%\begin{figure*}[h!]
%	\vspace{-.2cm}
%	\centering    
%	\begin{tabular}{c c}		
%		\subfigure[An LSTM cell]{\label{fig:LSTM}\includegraphics[width=0.330\textwidth]{figs/LSTM2.pdf}} &
%		\subfigure[A traditional RNN]{\label{fig:RNN}\includegraphics[width=0.48\textwidth]{figs/RNN2.pdf}}
%	\end{tabular}
%	\caption{Recurrent neural network and LSTM (Reproduced~\cite{olah2020}).} %Long short-term memory 
%	\vspace{-.4cm}
%\end{figure*}

Neural networks are among the most notable ML techniques, and they learn from labeled data to deal with unlabeled data afterwards~\cite{nielsenneural},\cite{Svozil1997}. Thanks to these characteristics, they have been widely used for different tasks, such as pattern recognition~\cite{Bishop:1995:NNP:525960}, face and smile detection~\cite{7961718}, classification~\cite{DUONG2020105326}, forecasting~\cite{RePEc:eee:intfor:v:14:y:1998:i:1:p:35-62}, to name a few. Recurrent neural networks (RNNs)~\cite{DBLP:conf/aaai/AlemanyBPG19} are a family of neural networks that are specialized in dealing with sequence data, \eg time series. %Figure~\ref{fig:RNN} depicts a typical recurrent neural network, where there are an input $x_t$, an output $h_t$ and a processing unit A. 
An RNN attempts to store information about past events and the output of a loop is provided as input for the next loop. A main drawback of this type of network is that it cannot learn well long-term dependencies. Thus, long short-term memory recurrent neural networks (or LSTMs for short) have been proposed to transcend the limitation~\cite{10.1162/neco.1997.9.8.1735}. An LSTM learns better long term dependencies by memorizing the input sequence of data. It has a mechanism to remove or add information to the cell state. In this way, LSTMs can remember dependencies for a longer time, %The way we update the hidden state and propagate information. 
as they are capable of retaining worthy/valuable information and discarding useless information by means of the \emph{sigmoid} and the \emph{tanh} functions. Two states are propagated to the next cell, \ie the cell state and the hidden state. The output of the previous unit together with the current input is fed as the input data for a cell.

\begin{figure}[h!]
	\centering
	\vspace{-.4cm}
	\includegraphics[width=0.680\linewidth]{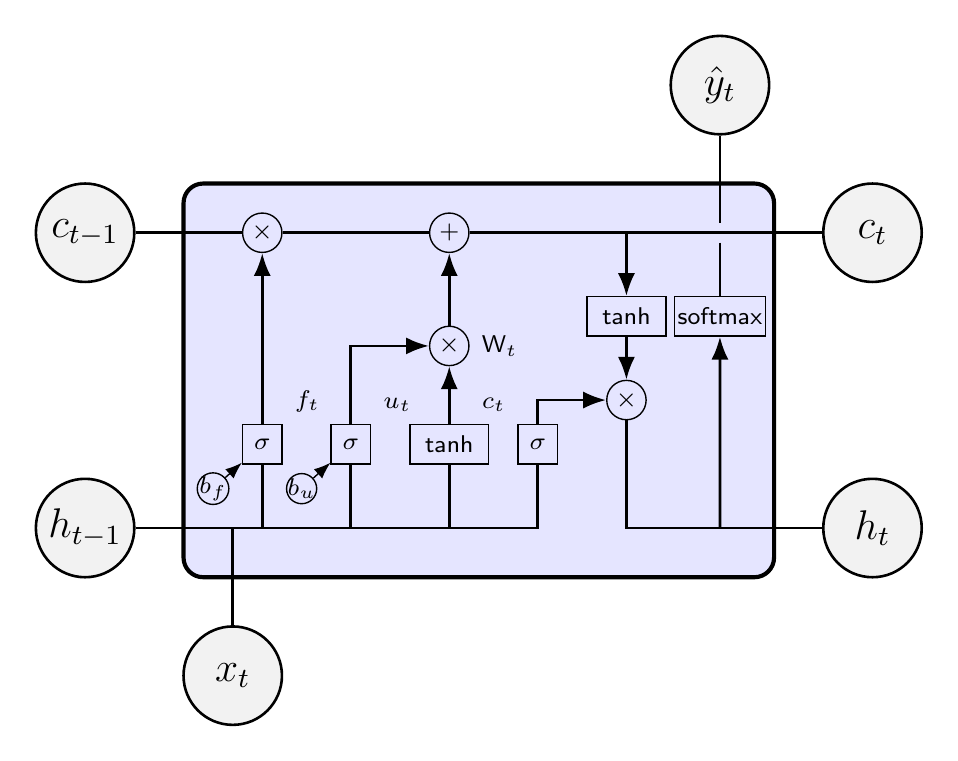}
	\vspace{-.2cm}
	\caption{The structure of an LSTM cell.} %~\cite{olah2020}
	\vspace{-.3cm}
	\label{fig:LSTM}
\end{figure}

In Fig.~\ref{fig:LSTM}, we depict the internal structure of an LSTM cell~\cite{olah2020}: %, we refer to %Fig.~\ref{fig:RNN} and 
%Fig.~\ref{fig:LSTM}: the LSTM cell 
there are various processing units, and each of them has its own functionalities. First, the sigmoid function is defined below.
\vspace{-.1cm}
\begin{equation} \label{eqn:sigmoid}
\sigma ( x) = (1 + exp(-x))^{-1}%\frac{1}{1 + exp(-x)}
\end{equation}
%Moreover, the following formulas are derived f
%From Fig.~\ref{fig:LSTM}, t
The following formulas are derived:% and they are given below:
\vspace{-.2cm}
\begin{equation} \label{eqn:forget}
f_t = \sigma (W_f[h_{t-1},x_{t}]+b_{f})
\end{equation}
\vspace{-.6cm}
\begin{equation} \label{eqn:update}
u_t = \sigma (W_{u}[h_{t-1},x_{t}]+b_{u})
\end{equation}
\vspace{-.6cm}
\begin{equation}
c_t = tanh (W_{c}[h_{t-1},x_{t}]+b_{c})
\end{equation}
\vspace{-.6cm}
\begin{equation}
W_{t} = c_{t-1} \cdot f_{t} + N_{t}.u_{t}
\end{equation}

%\begin{equation}
%C_{t} = C_{t-1} \cdot f_{t} + N_{t}.i_{t}
%\end{equation}
\vspace{-.2cm}
where $W$ and $b$ are the weight and bias matrices for different network entries. We compute the forget and the update values by means of Eq.~\ref{eqn:forget} and Eq.~\ref{eqn:update}, respectively. $h_{t}$ is the output of the current step and it is fed as the hidden state to the succeeding cell~\cite{9013316,Shi2019}.

%\begin{table*}[h!]
%	\footnotesize
%	\centering
%	\begin{tabular}{|c|c|} \cline{1-2}
%		Basic RNN  &  LSTM \\  \cline{1-2}				
%		\begin{tabular}{c}
%			%			\vspace*{.4cm}
%			\( \sigma ( x) = \frac{1}{1 + exp(-x)} \label{eqn:sigmoid2} \)     \\\\
%			\( sim_\beta(d,e)=\frac{|\mathbb{F}(d)\bigcap \mathbb{F}(e)|}{|\mathbb{F}(d)\bigcup \mathbb{F}(e)|}  \)  
%		\end{tabular}&				
%		\begin{tabular}{c}
%			%			\vspace*{.5cm}			
%			\( R_{e,i,p}=\frac{\sum_{q \in topsim(p)}r_{e,i,q} \cdot sim_{\alpha}(p,q)}{\sum_{q \in topsim(p)}sim_{\alpha}(p,q)}   \) \\\\ 
%			\( r_{d,i,p} = \overline{r}_{d} + \frac{\sum_{e \in topsim(d)}(R_{e,i,p}-\overline{r}_{e}) \cdot sim_{\beta}(d,e)}{\sum_{e \in topsim(d)}sim_{\beta}(d,e)}  \) \\\\			
%		\end{tabular}		
%		\\ \cline{1-2}
%	\end{tabular}	
%	%	\scriptsize
%	\caption{Compute similarities and predictions: $sim_{\alpha}(d,e)$ is the similarity between projects $p$ and $q$; $sim_{\beta}(d,e)$ is the similarity between declarations $d$ and $e$; $\mathbb{F}(d)$ and $\mathbb{F}(e)$ are the sets of invocations of $d$ and $e$, respectively.}
%	\label{tab:Formulas}
%	\vspace{-.7cm}
%\end{table*}

%$\rhd$~\emph{Softmax}: 

%The last fully-connected layer of an LSTM neural network normally uses 
%In classification, 
Softmax is used as the activation function to transform a set of real numbers to probabilities, summing to 1.0~\cite{10.1162/neco_a_00990}. %. of a class
For a set of C classes, we call $y_{k}$ as the output of the k$^{th}$ neuron, corresponding to the k$^{th}$ class. The final prediction is associated with the class getting the maximum probability, \ie $\hat{y} = argmax\medspace p_{k}$, $k \in 1..C$, where $p_{k}$ is defined as follows. %~%as specified   probability distribution
\vspace{-.1cm}
\begin{equation} \label{eqn:Softmax}
p_{k} = \frac{exp(y_{k})}{\sum_{k=1}^{C} exp(y_{k})}
%f_{om}(x,y) = (f_{im}*k)(x,y) = \sum_{i}\sum_{j}f_{im}(i,j).k(x-i,y-j)
\end{equation}

%LSTMs have been widely used to deal with sequential data, like predicting hurricane trajectories~\cite{DBLP:conf/aaai/AlemanyBPG19} or forecasting flood~\cite{Le_2019}. 

We propose a practical solution to the prediction of bus transit feed classification by training an LSTM, using data collected from different journeys. The next section presents our conceived approach in detail.

%The succeeding section brings in our approach and the architecture to realize a system providing transit feed specification.

%to predict future trajectories of a 

\section{Proposed Approach}\label{sec:approach}
%\todo[inline]{Ludovico+Phuong}

%Having a GTFS available  on a web server can drive the process to publish it as available for use of third parties. 
%Currently multiple software managing the bus schedule of an agency provides an additional feature of exporting the GTFS of the stored data. 

%In this section, we attempt to tackle the issue described in the previous sections: 

If an export feature to GTFS format is not available, then two scenarios are possible to enable the use of OTP: \emph{(i)} implement the missing export feature; \emph{(ii)} reverse engineering of the real time data tracking. Since the former is not a unique process and it has to be a customized feature for every software, with additional costs, we adopt the second scenario with an approach based on Machine Learning techniques to retrieve the transit graph.

Real-time tracking of fleets is a functionality based on a wide range of technologies and communication systems, \ie GPRS, GSM, GPS, microcontrollers, \etc used to track the position of a vehicle and eventually report it on a map. When a bus is equipped with these devices, it can immediately transmit to a server its current positions, that are then properly exploited.
%the final intent.  

%<<<<<<< HEAD
%\input{src/LSTM}
%
%
%\input{src/architecture}

%=======
%\subsection{Architecture}

%\subsection{Predicting bus trajectories} \label{sec:Methodology}

%Given the dataset, %a certain number of entries is selected as input, and the next entry is the label. 
%We exploited the dataset to perform 
%move this part later

\begin{figure}[h!]
	\begin{center}
		\includegraphics[width=\columnwidth]{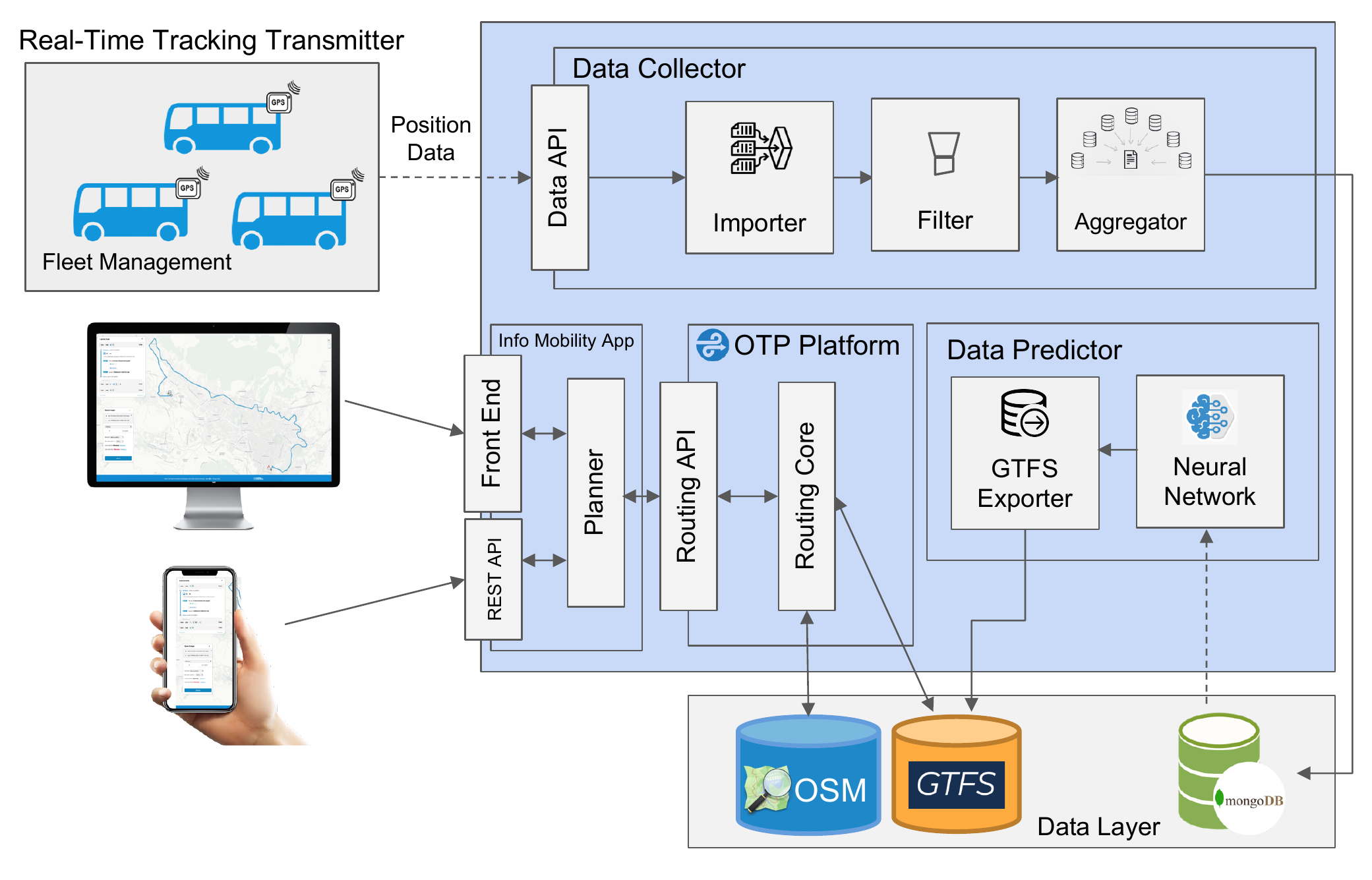}
		\caption{Overview of the info-mobility architecture.}
		\label{fig:infomobility-architecture}
	\end{center}
	\vspace{-0.5cm}
\end{figure}

%\subsection{System Architecture} \label{sec:Architecture}
We propose a practical solution to support infomobility systems with an automated process based on data mining and machine learning techniques~\cite{Garg:18} to deal with the lack of public transport information. The proposed architecture is depicted in Fig.~\ref{fig:infomobility-architecture}, and consists of various components: \textit{Real-Time Tracking Transmitter}, \textit{Data Collector}, \textit{Data Predictor}, \textit{Open Trip Planner (OTP) Platform}, and \textit{Info Mobility App}, whose functionalities are described %in detail 
below.

\texttt{Real-Time Tracking Transmitter.} Every bus of the fleet is equipped with a GPS real-time tracker, which regularly sends the collected data following a predefined configuration. The mounted device is a modular plug and play vehicle gateway supplying to fleet managers real-time visibility to location, driver's behavior, CAN Bus Interface, temperature and other engine data not relevant for the purposes of this work. Table~\ref{tab:Datasets} depicts an excerpt of the data collected \emph{in situ}, which is a tuple of five fields. In particular, %each entry is identified by an \emph{ObjectID}; %The data collected format can be summarized as following: 
%\textit{UnitID} is the identifier of the device installed on the vehicle; 
\textit{Latitude} and \textit{Longitude} are the lateral and longitudinal position of the bus at the time identified by \textit{Datetime}; \textit{Speed} is the corresponding velocity, and finally \emph{UnitID} is the bus's device identifier. Each device is configured with running parameters like the interval for sending data, type of data, and some additional ones. 

\texttt{Data Collector.} The collected data is then transmitted by the device through \emph{Data REST API}, and received by \emph{Data Collector}, which consists of the following three sub-components: \emph{Importer}, that receives the data from \emph{Data REST API}; \emph{Filter}, that performs various refinement steps to clean and eventually align the data, %can discard data according to some criteria (e.g., data are not well-formatted); 
and \emph{Aggregator}, that converts and saves the data in the prescribed format. %accordingly to the database structure. 
The data layer is delegated to a NoSQL database, MongoDB.\footnote{\url{https://www.mongodb.com/}} These types of databases are particularly suitable for representing graphs in public transport systems. 

% represents the date of the detection
%set considered in our evaluation
% installed

%and through the Neural Network
%\vspace{-.1cm} ^{k}

\begin{table}[t!]
	\small
%	\color{blue}
	\centering		
	\scriptsize
	\vspace{-0.2cm}
	\caption{An excerpt of the dataset.} % used in the paper
	\begin{tabular}{|p{2.0cm}|p{2.0cm}|p{0.5cm}|p{0.6cm}|p{1.2cm}|}	\hline
		\rowcolor{verylightgray}
		\textbf{Latitude} & \textbf{Longitude} & \textbf{Speed} & \textbf{UnitID} & \textbf{Time} \\ \hline %\hline		
		42.3724250793457 & 13.283947944641113 & 17 & 844852 & 2019-12-04 13:54:17  \\ \hline		
		42.372474670410156 & 13.284528732299805 & 13 & 844852 & 2019-12-04 13:54:22  \\ \hline		
		42.37250518798828 & 13.28501033782959 & 18 & 844852 & 2019-12-04 13:54:27  \\ \hline	
		42.372554779052734 & 13.285616874694824 & 19 & 844852 & 2019-12-04 13:54:32  \\ \hline
		%		42.37257385253906 & 13.286133766174316 & 16 & 844852 & 2019-12-04 13:54:37  \\ \hline
		%		6 & 42.37269973754883 & 13.28672981262207 & 19 & 844852 & 2019-12-04 13:54:42  \\ \hline
		%		7 & 42.37279510498047 & 13.287284851074219 & 17 & 844852 & 2019-12-04 13:54:47  \\ \hline
		%		8 & 42.37299728393555 & 13.28785514831543 & 19 & 844852 & 2019-12-04 13:54:52   \\ \hline
		%		9 & 42.37311935424805 & 13.288171768188477 & 3 & 844852 & 2019-12-04 13:54:57   \\ \hline
		%		10 & 42.37313461303711 & 13.288296699523926 & 10 & 844852 & 2019-12-04 13:55:02  \\ \hline
%		42.37269973754883 & 13.28672981262207 & 12 & 844852 & 2019-12-04 13:54:42  \\ \hline		
		\multicolumn{5}{c}{...} \\ \hline
		42.35724639892578 & 13.35239315032959 & 15 & 081102 & 2020-09-07 00:41.00 \\ \hline
		42.35682678222656 & 13.35276985168457 & 16 & 081108 & 2020-09-07 00:46.00 \\ \hline
		42.35724639892578 & 13.35239315032959 & 16 & 081102 & 2020-09-07 00:46.00 \\ \hline
		42.35682678222656 & 13.35276985168457 & 14 & 081108 & 2020-09-07 00:51.00 \\ \hline
%		42.35724639892578 & 13.35239315032959 & 20 & 081102 & 2020-09-07 00:51.00 \\ \hline
		\multicolumn{5}{c}{...} \\ \hline		
		42.360660552978516 & 13.351606369018555 & 43 & 081123 & 2020-10-01 11:27:09 \\ \hline
		42.325374603271484 & 13.398246765136719 & 33 & 081124 & 2020-10-01 11:27:09 \\ \hline
		42.35531234741211 & 13.403565406799316 & 3 & 081109 & 2020-10-01  11:27:09 \\ \hline
		42.371707916259766 & 13.347664833068848 & 25 & 081103 & 2020-10-01 11:27:11 \\ \hline
%		42.360626220703125 & 13.377833366394043 & 3 & 081110 & 2020-10-01 11:27:12 \\ \hline
%		42.35737228393555 & 13.352156639099121 & 0 & 081108 & 2020-10-01 11:27:11 \\ \hline
%		42.34553146362305 & 13.410273551940918 & 25 & 081125 & 2020-10-01 11:27:14 \\ \hline
%		42.32510757446289 & 13.399238586425781 & 33 & 081124 & 2020-10-01 11:27:14 \\ \hline		
		%		42.35682678222656 & 13.35276985168457 & 19 & 081108 & 2020-09-07 00:56.00 \\ \hline
		%		42.35724639892578 & 13.35239315032959 & 19 & 081102 & 2020-09-07 00:56:00 \\ \hline
	\end{tabular}
	\label{tab:Datasets}
%	\color{black}
	\vspace{-.2cm}
\end{table}

\texttt{Data Predictor.} Once the data has been properly curated and aggregated, the \emph{Data Predictor} sub-component periodically selects it from MongoDB and provides it as input for the \emph{LSTM} Neural Network, which predicts the recurrent patterns of the buses based on the extracted data. The designed and implemented LSTM forecasts future trajectories of a bus, Y = $\{y_{t}\}$, $t \in T^{f}$, based on a set of features from recent past events, \ie X = $\{x_{t}\}$, $t \in T^{p}$, where $T^{p}$ and $T^{f}$ are time in the past and the future, respectively. The input data to feed the system is a tuple of the form $x_{t}$=$<$\emph{latitude, longitude, speed}$>$ which captures the current location and speed of a bus. In practice, given an accumulated series of recent bus locations, the system is expected to predict the bus' future whereabouts within a specific amount of time, including also the stops and timetable.
%{\bf what is the parameter k here? We should write a line to say where k is or will be specified later. In this informal discussion we might even replace it}
%{\bf here there is two times GTFS exporter, maybe one occurrence is mistaken}
The predicted data is then provided as input to \emph{GTFS Exporter}, which detects routes, trips and bus stops, where trips are individual components of a route. %The last sub-component, \ie \emph{GTFS Exporter}, 
The sub-component converts the predicted features into the GTFS format, and stores them into the Data Layer. The feeds produced by Exporter are tested using FeedValidator, a command line tool that extracts a GTFS feed to provide an HTML report.\footnote{\url{https://github.com/google/transitfeed/wiki/FeedValidator}} The network's parameters are shown in Table~\ref{tab:Configurations}.  % \cite{9013316}

%In Fig.~\ref{fig:Network}, we illustrate the LSTM network architecture~\cite{olah2020} to realize the \emph{Data Predictor} sub-component, and 

%\begin{figure}[h!]
%	\begin{center}
%		\vspace{-.3cm}
%		\includegraphics[width=0.80\columnwidth]{figs/Network.pdf}
%		\vspace{-.3cm}
%		\caption{The network architecture.}
%		\label{fig:Network}
%	\end{center}
%	\vspace{-.4cm}
%\end{figure}

\begin{table}[h!]
	\small
%	\color{blue}
	\centering		
	\scriptsize
	\footnotesize
	\vspace{-0.2cm}
	\caption{Network's configurations.} 
	\begin{tabular}{|p{3.0cm}|p{1.5cm}|}	\hline
		\rowcolor{verylightgray}
		\textbf{Parameter} & \textbf{Value}   \\ \hline 
		Batch size & 30    \\ \hline		
	    Number of hidden units & 400  \\ \hline		
		Learning rate & 5$\times$10$^{-4}$  \\ \hline	
		Activation funtion & ReLU  \\ \hline	
		Number input features & 3  \\ \hline
		Number output features & 3  \\ \hline
%		 &   \\ \hline		
	\end{tabular}
	\label{tab:Configurations}
%	\color{black}
	\vspace{-.3cm}
\end{table}

%Currently, since no useful data from public administrations or transport companies are available concerning bus stops location and routes, covering the urban territory, the outputs of these algorithms will allow to extract information from real time positions of bus stops (associated with the route considered), the arrival and waiting times of a bus. 
%Finally, 

\texttt{OTP Platform.} The data is ported to the OTP platform by means of a REST API, \ie \textit{Routing API}. As anticipated, OTP is a cross-platform multimodal route planner, which combines the OSM and GTFS data for supporting public transport routing, \ie the \textit{Routing Core} component. 

\texttt{Infomobility App.} Finally, the \textit{Info Mobility App} component provides a map, through a Web site or a mobile app, and includes a travel planner that allows users to create suitable itineraries. %facilitate the creation of itineraries by citizens and tourists.

The open data collected by the system can be integrated and used by public administrations and transport companies to improve their services or to schedule the maintenance of fleet or road infrastructures. This architecture represents just a prototypical implementation that we are going to extend and update with the specific components after the design of the real system. 

%It is worth noting that with our proposed approach,  all the collected GPS data does not necessarily need to be stored in a database, as it is the case with the static program mentioned in Section~\ref{sec:MotivatingExample}. Once the GPS data has been used to train the system, it can be discarded to give place to new incoming training data. In this way, we see an additional advantage of the ML system: the knowledge learned and extracted from data is memorized in the internal neural network of parameters, including weights and biases, without resorting to any database, making the Data Collector component optional.

%In the next section, we describe our pilot study, that involves a bus serving more than one route, together with a preliminary evaluation of the prototype, based in the dataset collected \emph{in situ}.

\section{Evaluation}\label{sec:evaluation}
%\todo[inline]{Phuong +Ludo: Compare the retrieved data with ML with existing routes in GTFS provided by AMA}

%In the evaluation, we made use of the Fruits-360 dataset~\cite{article} and a recent implementation\footnote{\url{https://github.com/rwightman/gen-efficientnet-pytorch}} of \EN and \MN which was built on top of the PyTorch framework.\footnote{\url{https://pytorch.org}} Furthermore, we investigate whether imposing transfer learning is beneficial to the given dataset by adopting pre-trained weights from the ImageNet dataset~\cite{Russakovsky:2015:ILS:2846547.2846559}. 

%This section describes the dataset used. 
This section explains in detail the methods and materials used to evaluate the proposed approach on a real-world dataset. %as well as the experimental results. 
In particular, Section~\ref{sec:ResearchQuestions} introduces the research questions raised to study the approach's performance. Section~\ref{sec:Dataset} gives an overview of the dataset used in the evaluation, which contains real GPS data collected from buses. %The metrics utilized to measure the system's performance are presented in Section~\ref{sec:Metrics}.

% exploited in the evaluation

%, while Section~\ref{sec:Methodology} describes the experimental settings

\vspace{-.2cm}
\subsection{Research questions} \label{sec:ResearchQuestions}
%To better understand the evaluation we planned, we can summarize the research questions as:
%We consider 
%The following research questions have been considered to study 
The applicability of our proposed approach has been studied by means of the following three research questions:

\vspace{.1cm}
\noindent
$\rhd$~\rqfirst Using the real dataset collected from GPS trackers, we evaluate if our approach is able to predict the future location of a bus, given its previous locations, aiming at reconstructing the entire route. %of service. 

\vspace{.1cm}
\noindent
$\rhd$~\rqsecond It is necessary to forecast and provide commuters with the next bus stop(s) while they are en route. We evaluate if the approach can early notify bus stops by means of the labeled data collected from the local travel agency. 

% with respect to a series of recent locations of a bus
%, the agency managing the public transport system of L'Aquila

\vspace{.1cm}
\noindent
$\rhd$~\rqthird We are interested in understanding if the conceived framework is able to provide instant recommendations while commuters are travelling. This is important in practice, since delay in producing predictions may induce inaccuracy. To measure the prediction time, we made use of a laptop with Intel Core i5-7200U CPU@2.50GHz$\times$4, 8GB RAM, and Ubuntu 16.04 as the operating system. 

%\begin{itemize}
%    \item[RQ1] Are we able to retrieve the bus route from the GPS traces?
%    \item[RQ2] Can the prediction approach be used to deduce missing information?
%\end{itemize}

\vspace{-.2cm}
\subsection{Data Extraction and Cleaning} \label{sec:Dataset}
%For the evaluation of
%We reconstructed the OTP graph from Excel files provided by 

%Index is: 909213
%The number of buses: 10
%006006081108	187400
%006006081109	13360
%006006081110	39088
%844852	157242
%006006081123	795
%006006081102	162262
%006006081124	53425
%006006081125	59019
%006006081103	53962
%006006081107	182661

\begin{figure}[h!]
	\centering
	\vspace{-.4cm}
	\includegraphics[width=0.50\linewidth]{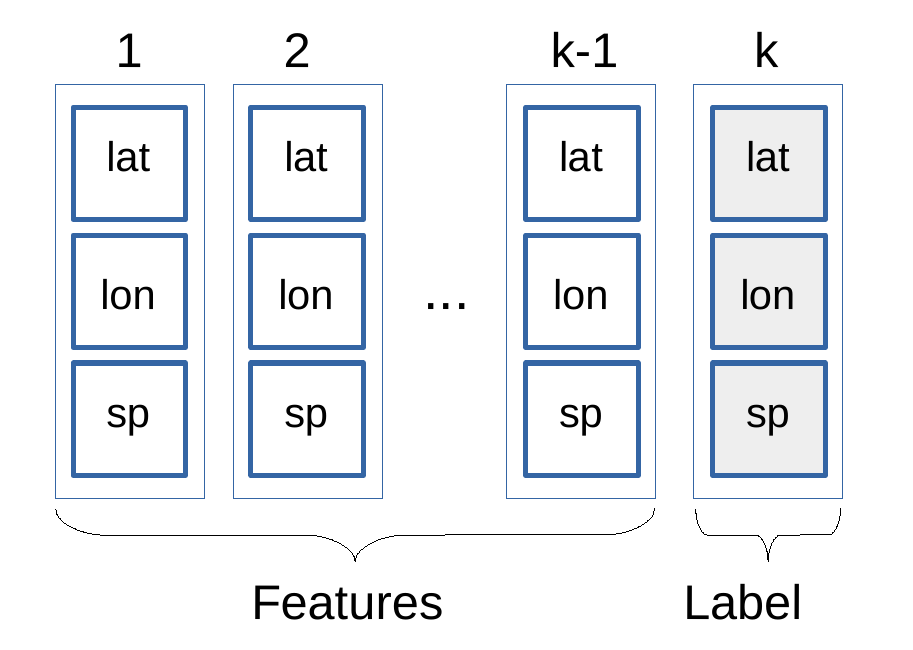}
	\vspace{-.2cm}
	\caption{Extraction of feature and label.}
	\vspace{-.2cm}
	\label{fig:Setting}
\end{figure}

To evaluate the proposed approach, we made use of a set of GPS data records collected by a fleet consisting of ten buses of the local travel agency (AMA), from December 2019 to October 2020. The dataset was provided in a CSV file, an excerpt of which with rows in ascending order of time is shown in Table~\ref{tab:Datasets}. There are 1,704,763 rows and five columns, whose meaning has been explained in Section~\ref{sec:approach}. %By carefully inspecting the dataset, we noticed that given a specific bus, there are several consecutive rows with exactly the same locations and speed, and we suppose that these are noise generated by the GPS devices. Thus, we performed some pre-processing steps to remove such rows.} %infrastructured equipped with a GPS tracker of which we manually built the OTP graph. 
%The prediction is 
Starting from the original CSV file, we performed some filtering steps to clean the data. For instance, we detected several rows with distinct Latitude and Longitude coordinates, but their corresponding Speed is equal to 0. This means that the bus was still moving despite the claimed zero velocity. Or there are several consecutive rows with exactly the same locations and speed. Thus, we assumed that these rows are noise caused by the GPS device and needed to be removed. Afterwards, the dataset was sampled as shown in Fig.~\ref{fig:Setting} to create the feature sets. A data unit is as a tuple of the form $<$\emph{lat, lon, sp}$>$, which represents the latitude (\emph{lat}) and longitude (\emph{lon}) coordinates, and the current speed (\emph{sp}). A block of data consists of consecutive $k$ data units, where the first $k$-$1$  units are the features, while the $k^{th}$ entry is used as the label. The parameter $k$ can be configured during the evaluation. %A set of such k entries forms as a block, 
In this way, the original dataset is divided into several blocks. This division aims to simulate the proposed approach in a real deployment scenario: given a bus, the system predicts its next position based on the most recent locations.

%to provide input for the LSTM network as follows.
%as follows. In practice, it means that 

%To study the approach performance on 
%Being trained with the training data, the system is expected to predict the correct future location, \ie the ground-truth label in the testing data. 
%We use k=5 consecutive entries (cf. Fig.~\ref{fig:Setting}). 
%We create blocks of k consecutive entries, and assign a unique identifier for a certain number of block, called groups.

From the dataset, we extracted 60\% of the blocks as training data, 20\% as validation, and the remaining 20\% as testing data.  
Eventually, the extraction resulted in a set of 12,000 blocks as training data, and 4,000 blocks for both the validation and the testing set. While the testing data is used to evaluate the overall performance, we exploited the validation data to calibrate the model, \eg for choosing a suitable set of parameters, such as learning rate or network size. The next section reports on the outcomes we got from our empirical study. %that need to be predicted, and stored as ground-truth data. 

% (Italy)

%
%\subsection{Metrics} \label{sec:Metrics}
%
%Similar to different related studies, in this work, the following metrics are used to compare the predicted locations with the real ones.
%
%\vspace{.1cm}
%\noindent
%$\rhd$~Average Displacement Error (ADE): the metric computes the mean euclidean distance.
%
%\begin{equation}
%ADE= \frac{ \sum_{t}^{b} \left \| Y_t - \hat Y_t \right \| }{T_{pred}-T_{obs} }
%\end{equation}

%\vspace{.1cm}
%\noindent
%$\rhd$~Final Displacement Error (FDE).

\vspace{-.2cm}
\section{Results and Discussions} \label{sec:results}

%Discuss the results here.

This section presents and analyzes the results obtained through the experiments, by answering the research questions in Sections~\ref{sec:RQ1},~\ref{sec:RQ2}, and~\ref{sec:RQ3}. Afterwards, we provide some discussions on the conceived framework.

\vspace{-.2cm}
\subsection{\rqfirst} \label{sec:RQ1}

%In total, there are. 

\begin{figure*}[h!]
	\vspace{-.2cm}
	\centering    
	%	\color{blue}
	\begin{tabular}{c c}
		\subfigure[Learning rate $\gamma$ = 1 $\times$ 10$^{-4}$]{\label{fig:TrainLoss1}\includegraphics[width=0.40\linewidth]{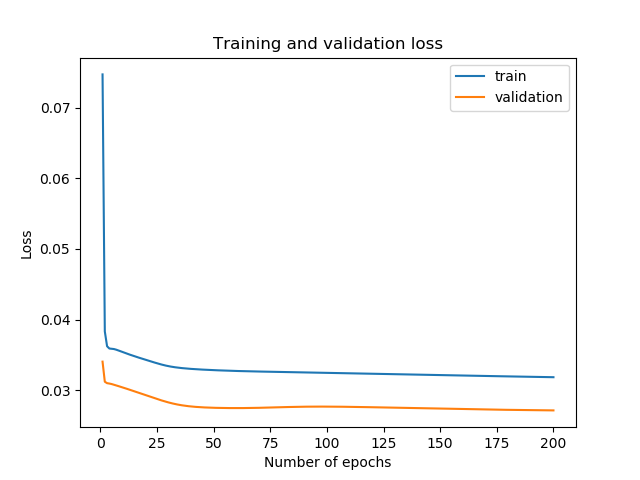}}  &
		\subfigure[Learning rate $\gamma$ = 5 $\times$ 10$^{-4}$]{\label{fig:TrainLoss2}\includegraphics[width=0.41\linewidth]{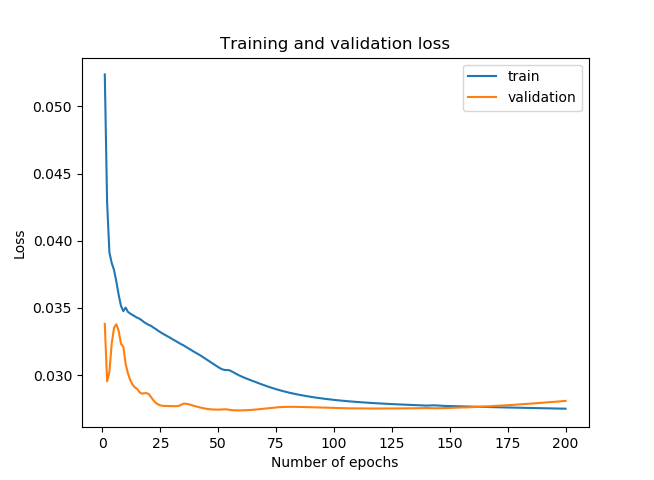}} 
	\end{tabular}
	\caption{Training and validation loss.}
	\vspace{-.2cm}
	%	\color{black}
\end{figure*}

An important issue in ML is the applicability of a model, \ie the probability that it works %in the field, 
with real data. If a system once trained fails to perform well even on the training data, it is considered to be underfitting. In contrast, if it learns effectively on the training data but performs poorly on the validation data, then it becomes overfitting. To investigate if the model can be \emph{good fit}, \ie it is neither underfitting nor overfitting, during the training process we exploited the validation data (\ie 20\% of the dataset) to observe the training and validation loss.
%following experiment. 
%{\bf I'm confused here, how does this relate with what previously said, that is 60 pc training, 20 validation and 20 test? And what an epoch is is still a bit unclear?}
%Starting from the data divided for the cross-validation evaluation, we further split each training fold %the original training data 
%into two parts: 80\% for training and 20\% for validation, and
Moreover, the number of epochs was set to 200, where an epoch is %a complete round of introducing all the training items to the network, \ie
an iteration of learning with all the training data. 
We also investigated the effect of the learning rate by using two values, \ie $\gamma=1\times10^{-4}$ and $\gamma=5\times10^{-4}$. The loss recorded during the experiments is shown in Fig.~\ref{fig:TrainLoss1} and Fig.~\ref{fig:TrainLoss2}, respectively. %for different value of $\gamma$. 
In Fig.~\ref{fig:TrainLoss1}, we see that there is always a gap between the training and validation loss, implying that the model is overfitting and cannot converge even after a large number of epochs. This demonstrates that the corresponding learning rate $\gamma=1\times10^{-4}$ is too small and does not induce an efficient training. Instead, in Fig.~\ref{fig:TrainLoss2}, a different pattern can be seen: after a dramatic fall, the loss curves start decreasing slowly once a threshold is reached, \eg after 75 epochs. The curves for training and validation loss stay close and they are nearly identical, which means %In this respect, we come to the conclusion 
that with %the learning rate 
$\gamma=5\times10^{-4}$ the model becomes less overfitting and has a quite good fit. By testing with other larger values, we realized that the model fits worse. Thus, we conclude %came to the conclusion 
that $\gamma=5\times10^{-4}$ is the most suitable learning rate for the model on the given dataset.

\begin{figure*}[h!]
	\vspace{-.2cm}
	\centering    
	%	\color{blue}
	\begin{tabular}{c c}
		\subfigure[Latitude]{\label{fig:LatitudePrediction}\includegraphics[width=0.38\linewidth]{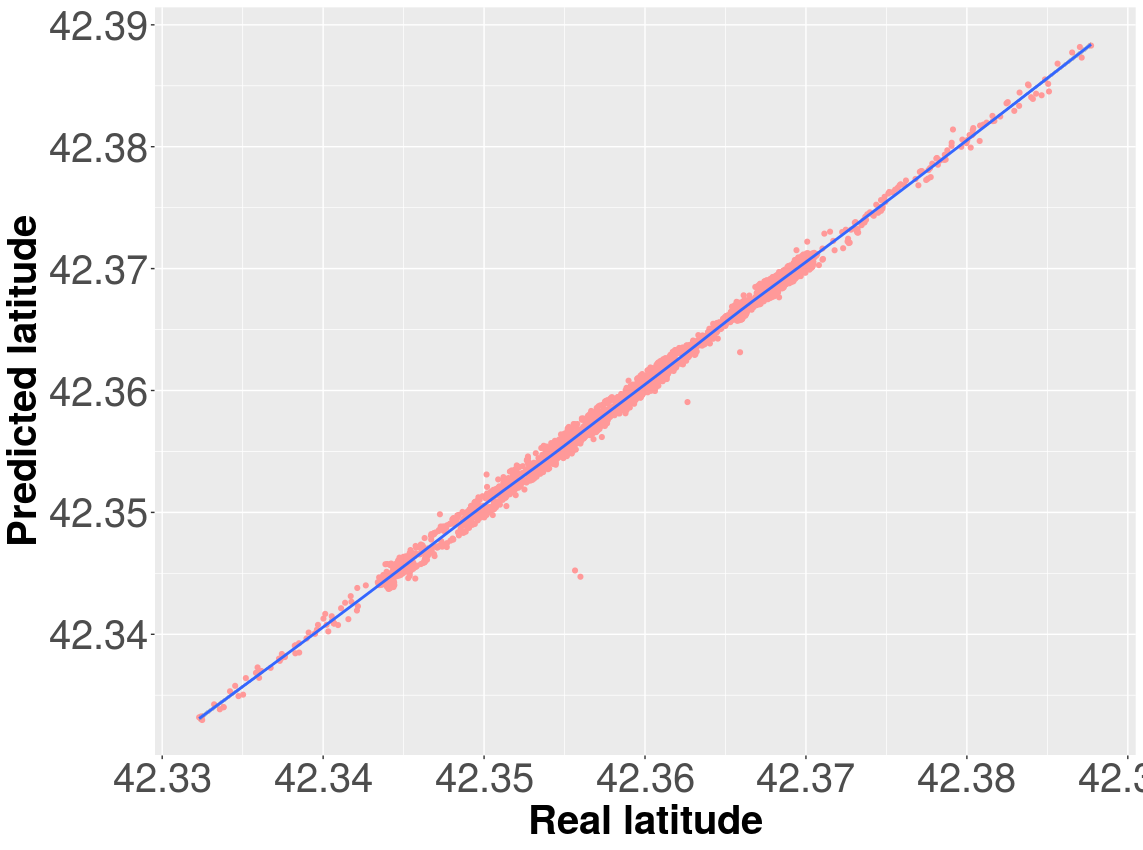}}  &
		\subfigure[Longitude]{\label{fig:LongitudePrediction}\includegraphics[width=0.38\linewidth]{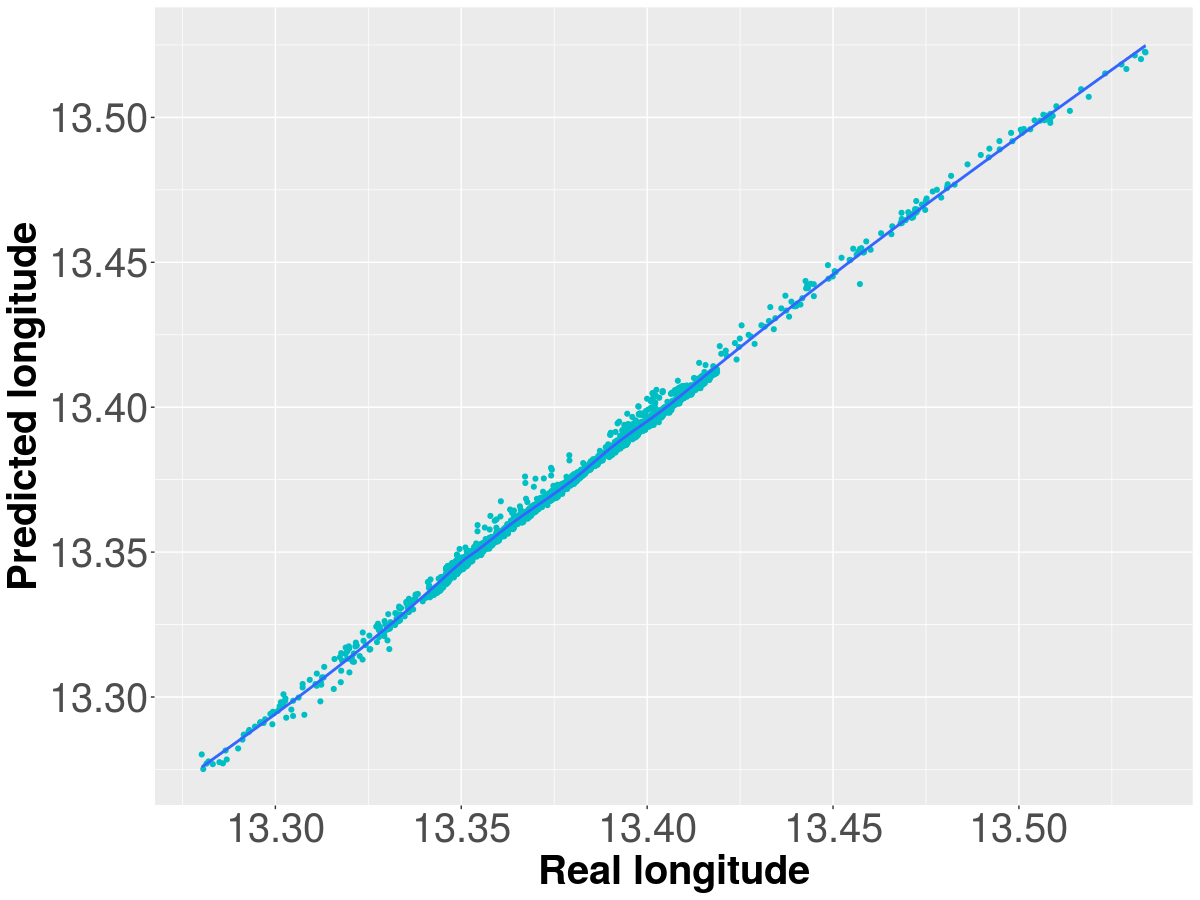}} 
	\end{tabular}
	\caption{Prediction of bus locations for ten buses.}
	\vspace{-.2cm}
	%	\color{black}
\end{figure*}

The relationship between the predicted latitude, longitude and their ground-truth ones for all the 4,000 data points corresponding to the testing set is depicted in Fig.~\ref{fig:LatitudePrediction} and Fig.~\ref{fig:LongitudePrediction}. In particular, the x-Axis and the y-Axis represent the real and the predicted coordinates, respectively. Both figures show that, despite some outliers, the lines representing the coordinates stay close to the 45\textdegree~diagonal, indicating that the predicted positions match well with the real ones. We computed the errors for latitude and longitude using the RMSE function~\cite{gmd-7-1247-2014}: a low RMSE value corresponds to a good prediction, and an RMSE equalling to 0 implies a perfect match. %{\bf what is this RMSE function? And what means having such values?} 
The final RMSE scores obtained for latitude and longitude are 5.2$\times10^{-4}$ and 2.6$\times10^{-4}$, respectively.

\begin{figure*}[h!]
	\centering
	\includegraphics[width=0.92\linewidth]{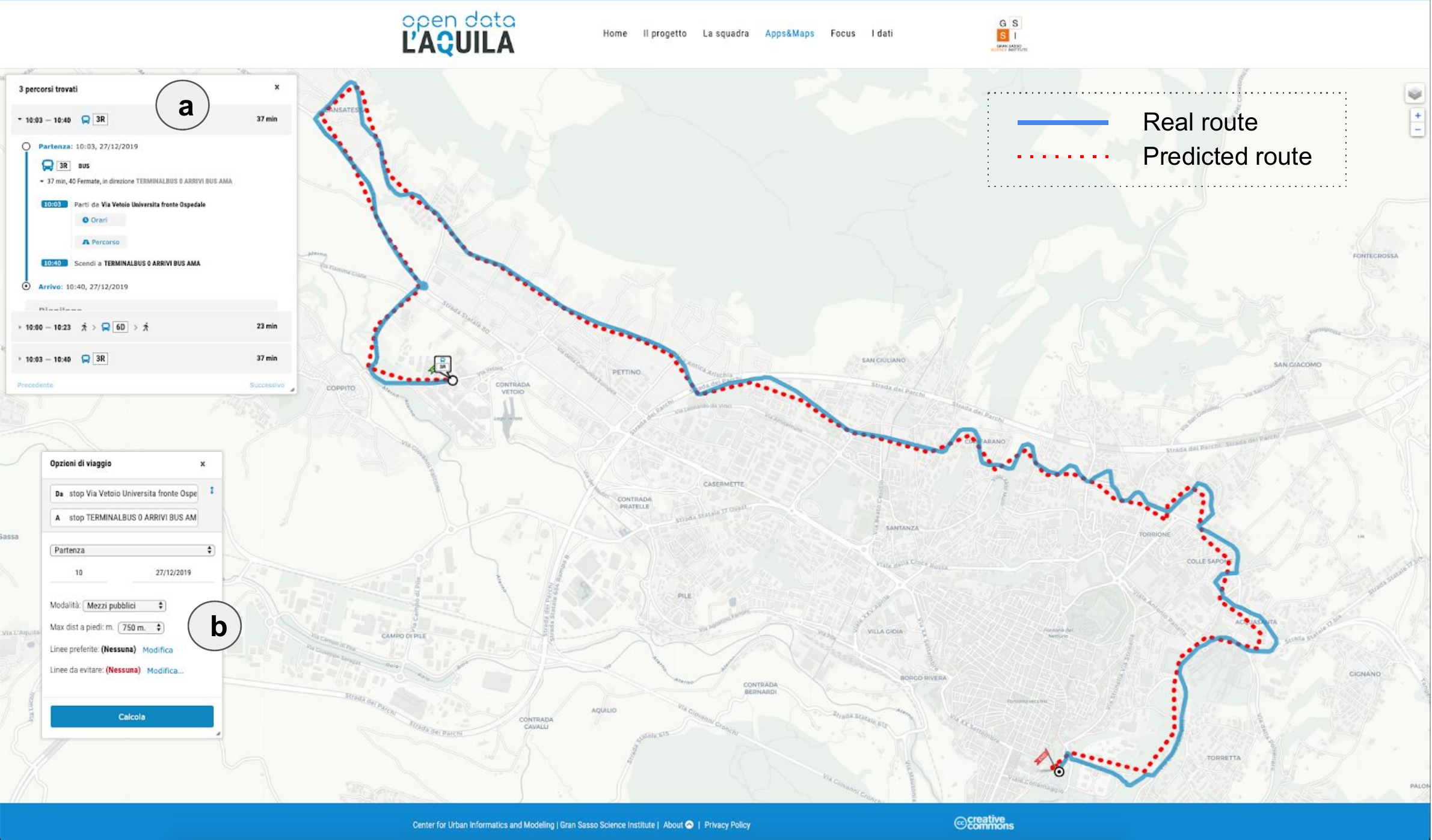}
	\caption{Prediction of bus routes.}% reported in OTP
	\label{fig:prediction-result-routes}
	\vspace{-.2cm}
\end{figure*}

Figure~\ref{fig:prediction-result-routes} shows the route obtained by the model (dotted red line) and the real one constructed from GPS data (continuous blue line). As it can be seen, the left upper parts of both lines match well with each other, implying that the LSTM obtains a good prediction performance, and this is consistent with the results presented in Fig.~\ref{fig:LatitudePrediction} and Fig.~\ref{fig:LongitudePrediction}. However, in the lower right parts there is a gap between the lines, corresponding to the cases where the model fails to correctly predict the route. We suppose that this inaccuracy is caused by noise in the collected data: as we mentioned in Section~\ref{sec:Dataset}, the coordinates reported by the GPS device are sometimes incorrect. GPS errors can occur for various reasons~\cite{tiwari2010appraisal}, \eg satellite related errors, receiver related errors. In this case, the more data is fed to the model, %the dataset increases in dimension 
the better the accuracy is. %sometimes reports wrong coordinates. 
%\MF{what is the meaning of this sentence?} \PN{I rephrased the sentence to say that the model fails if the input data is incorrect.}
Furthermore, the result is shown directly in the OTP system, as it can be seen from the label (\textsc{b}) offering the travel planner functionality and (\textsc{a}) giving the path of the required trip on the map. %Figure~\ref{fig:rq2} depicts a typical example of errors introduced by human activities: this route has been provided by the agency in order to test the approach as a shape file. The route is shifted far away from the correct one. 
By processing the real time tracking data, the model manages to predict the correct route %as it can be seen in the figure, 
fixing the wrong path in this specific case. Still, we were not able to filter out all possible errors, and this is the reason why the model here produces inaccurate routes and needs to be improved. We consider this as an interesting future work. 

\vspace{.2cm}
\begin{tcolorbox}[boxrule=0.86pt,left=0.3em, right=0.3em,top=0.1em, bottom=0.05em]
	\small{\textbf{Answer to RQ$_1$.} Using the available transit specification data, the proposed approach is able to reconstruct routes and trips.}
\end{tcolorbox}

%the dataset contains many rows with different coordinates, however the corresponding velocity is equal to 0.
%a clear pattern can be seen: T
%We see that while the training accuracy and validation accuracy are almost concordant with each other, the training loss and validation loss are not. A common pattern for the validation plots is
% Furthermore there is always a gap between the plot of validation loss and that of the training loss. In other words, the model appears to be slightly overfitting by the current configuration. We assume that this happens, again, due to the limited amount of data used for training~\cite{10.1162/neco_a_00990}. Thus, to further examine the model, we increased the amount of data for training by specifying the following ratio: 95\% of the data is used for training and only 5\% is used for validation. The resulted learning curves for this setting are depicted in Fig.~\ref{fig:TrainLoss1},~\ref{fig:TrainLoss2}. 

\vspace{-.2cm}
\subsection{\rqsecond} \label{sec:RQ2}

%This is important in practice since the system should be able to 

In practice, it is important to provide passengers with an estimation of the time left to the next stop while the bus is en route, together with its location. In this research question, we performed additional experiments to investigate if our proposed approach can be used to predict bus stops along the routes.

%For this purpose, 
We made use of a set of labeled data provided by the travel agency, also by cross checking with %crossing information with 
online locations. The data consists of a list of 300 bus stops with their latitude and longitude. These coordinates were injected into the dataset presented in Section~\ref{sec:Dataset} by adding one more column representing bus stops, \ie the column is set to 1 in the corresponding row for a stop, 0 otherwise. To train the model, we also employ the same data format shown in Fig.~\ref{fig:Setting}: we use $k$-$1$ consecutive tuples as input features, and the $k^{th}$ tuple corresponds to a bus stop. In this way, we simulate a real-world scenario: the model needs to predict a bus stop by taking into consideration its previous locations in the route.

\begin{figure}[h!]
	\begin{center}
		\vspace{-.3cm}
		\includegraphics[width=0.8\columnwidth]{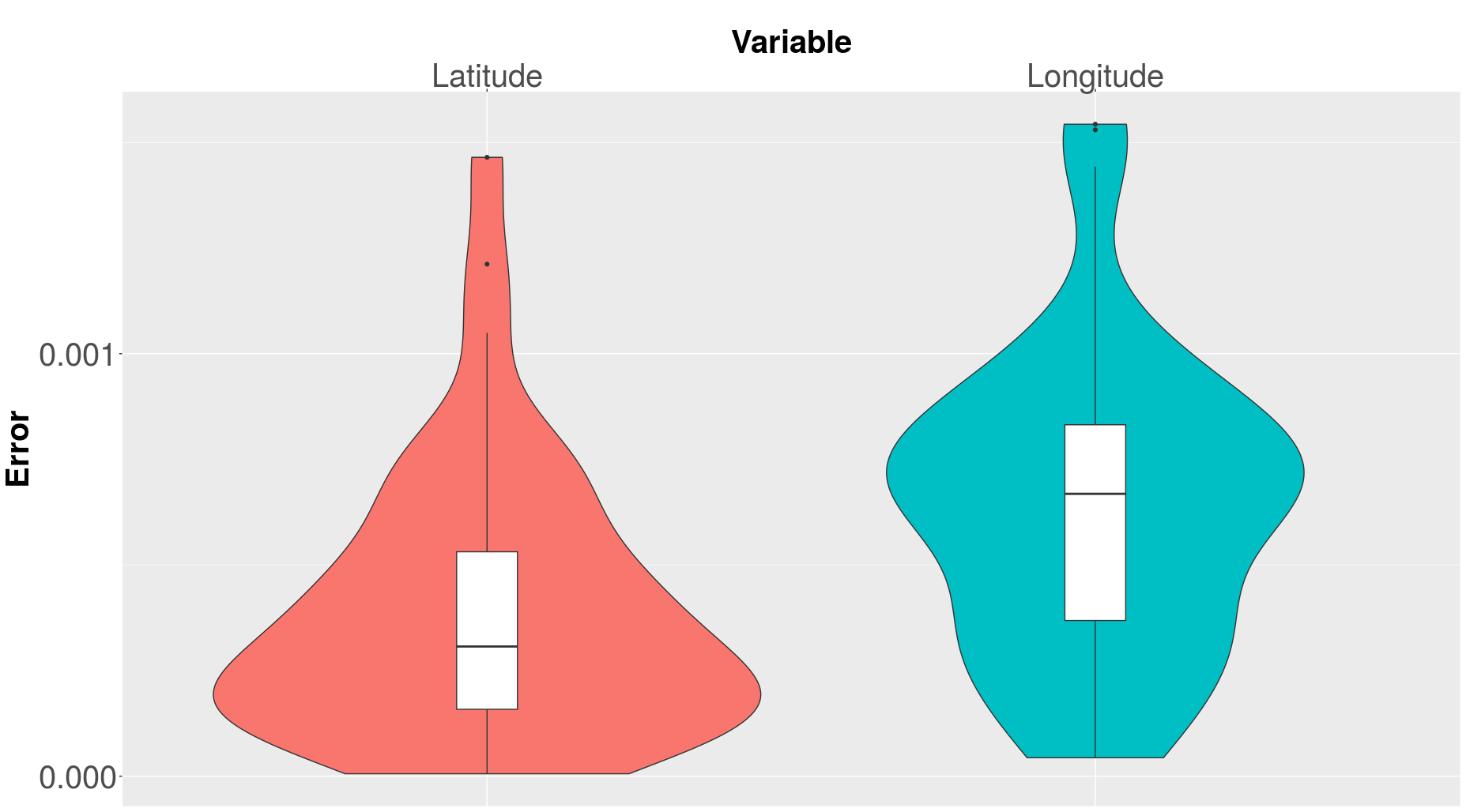}
		\caption{Prediction errors for bus stops.}
		\label{fig:error}
		\vspace{-.4cm}
	\end{center}	
\end{figure}

% We also trained the model using the same
%for all the testing data points. %Both figures show that the lines representing the coordinations stay close to the 45\textdegree~diagonal, indicating that the predicted positions match well with the real ones. %definitely this initial experiment is quite satisfying. %It can happen that in specific points the route is individuated with strange directions as in.

\begin{figure*}[h!]
	\begin{center}
		\includegraphics[width=.9\textwidth]{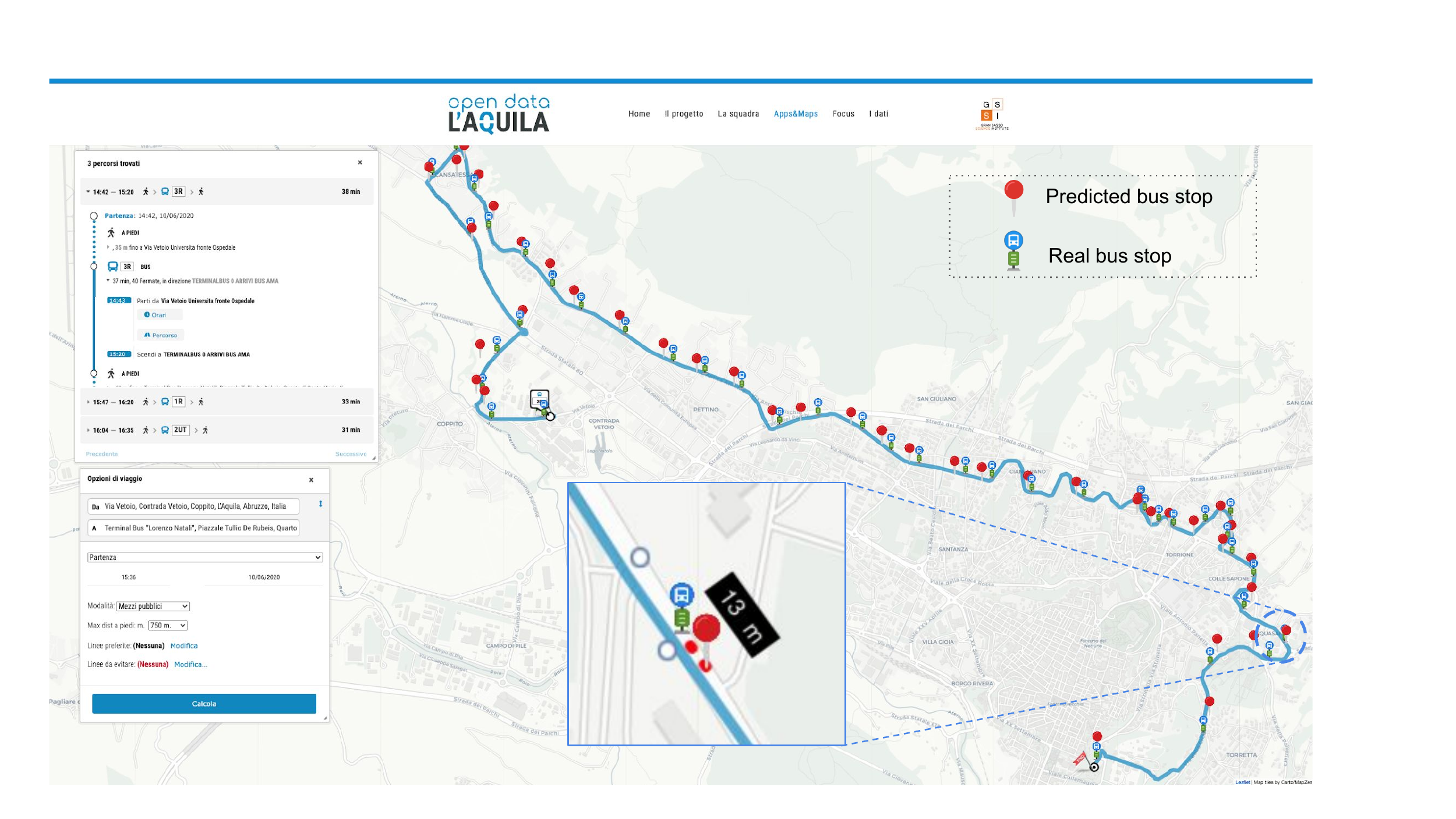}
		\caption{Prediction of bus stops.}
		\vspace{-.5cm}
		\label{fig:BusStops}
	\end{center}	
\end{figure*}

By running on the provided dataset, we obtain predicted locations for all the bus stops. We compute the errors between the predicted bus stops and the real ones and depict them in the violin boxplots in Fig.~\ref{fig:error}. The figure shows that most of the errors for latitude and longitude are lower than 1$\times$10$^{-3}$. Moreover, the computed error yields an RMSE of 7$\times$10$^{-4}$, implying that the predicted locations and the real ones are almost identical. We plot in Fig.~\ref{fig:BusStops} the outcomes of the predictions. It is evident that most of the predicted bus stops match well with the real ones. %As it can be seen in the figure, while our model can predict the bus stops, there are still certain errors among the predicted coordinates and the real ones, \ie 13m reported in the zoomed snapshot in Fig.~\ref{fig:BusStops}. 
%Still, while our model can predict the bus stops, there are certain errors, %among the predicted coordinates and the real ones, 
%\eg 13m reported in the zoomed snapshot in Fig.~\ref{fig:BusStops}. 
One of the limitations of the approach is the precision of the GPS position when GPS is not available, or when the signal is lost, but nevertheless this can be partially solved by fine-tuning the \emph{update time frequency} of the device; at the moment, according to our tuning, such a frequency has been fixed to 10 seconds.

The result confirms the usefulness of the approach, in cases when bus stops are part of the unavailable geo-tagged information. Applying this approach can easily help travel agencies to locate bus stops and store directly this information in infomobility systems, with a minimum amount of data, that can be quantified in a certain number of days of tracking, thanks to the obtained accuracy.

\vspace{.2cm}
\begin{tcolorbox}[boxrule=0.86pt,left=0.3em, right=0.3em,top=0.1em, bottom=0.05em]
	\small{\textbf{Answer to RQ$_2$.} Our approach can predict bus stops using the provided labeled data.}
\end{tcolorbox}

\vspace{-.2cm}
\subsection{\rqthird} \label{sec:RQ3}

%The core of our approach is an LSTM neural network, which has been trained with data collected from buses with mounted GPS devices. %More importantly, the proposed framework can process real-time data to provide commuters with updated information about the routes and trips. 
To validate the feasibility of our approach in practice, we measured the time needed for the framework to train and test on the dataset introduced in Section~\ref{sec:Dataset}. %The dataset consists of trajectories coming from 10 buses collected within four months by AMA, the local bus company of the city of L’Aquila. 
The testing data contains 15,600 locations spread across different routes. By running on an ordinary laptop using 200 epochs, we measured a duration of 80 minutes and 2 seconds for the training phase and testing phase, respectively. It is worth noting that in practice the most time consuming phase, i.e. the training one, can be done offline, %only once for a training set, 
for example overnight, and thus it does not affect the deployment/testing phase. In particular, on average the framework needs only 2/15,600 = 1.28$\times$10$^{-4}$ seconds to provide a recommendation for a location. In this sense, we conclude that the framework is suitable for working with real-time data, as it can generate instant recommendations while the bus is en route.

\vspace{.2cm}
\begin{tcolorbox}[boxrule=0.86pt,left=0.3em, right=0.3em,top=0.1em, bottom=0.05em]
	\small{\textbf{Answer to RQ$_3$.} While the training phase is time consuming, it can be done offline, only once for a training set. In summary, the proposed network is capable of providing real-time predictions.}
\end{tcolorbox}

%\subsection{\newcontent{Discussions}} \label{sec:Discussions}
%This section provides some discussions on the obtained results. 

\noindent \textbf{Discussion.} Our proposed framework has been built on top of an LSTM, being able to provide commuters with updated information about routes and stops while they are traveling. By means of an empirical evaluation, we validated the feasibility of the approach using a real-world dataset collected from buses equipped with GPS devices. In fact, without having data about bus trips of the fleet, real-time tracking services cannot be enabled, since the collected data cannot refer to a specific trip. The proposed approach is able to reconstruct the entire graph of a bus service, which is a pre-requirement for enabling real-time tracking. Compared to a static program that analyzes coordinates by connecting to a database populated with GPS data (cf. Section~\ref{sec:MotivatingExample}), our approach brings significant advantages. In particular, there is no need to store all the collected GPS in a storage layer, as it is the case with the static program. Once the GPS data has been used to train the system, it can be discarded to give place to new training data. In other words, the knowledge learned from data is memorized in the internal parameters of the neural network, %including weights and biases, 
without resorting to any database, making the Data Collector component optional. Moreover, it is worth noting that the file size needed to store the network's weights is much smaller than the GPS traces stored in a database, \eg hundreds of kilobytes compared to hundreds of megabytes. In this way, the framework is compact and it can run on lightweight devices, such as smartphones, facilitating portability. Furthermore, thanks to the internal trained weights, it works offline without the need to consult a central server, thus providing bus specifications even when the commuter is not connected to the network. Moreover, the proposed approach is resilient against noise resulting from inaccurate data, making the developed framework robust and the applications insensitive to small perturbations.

%In this sense, we need a practical solution to transcend the limitations. 

% Fig.~\ref{fig:limitation1}
%
%\begin{figure}[h!]
%	\begin{center}
%		\includegraphics[width=\columnwidth]{figs/limitation1.png}
%		\caption{Limitation to be discussed}
%		\label{fig:limitation1}
%	\end{center}
%	\vspace{-0.5cm}
%\end{figure}
%\input{src/threads}
\section{Conclusion}\label{sec:conclusione}

% the approach is better because we start from scratch
% the approach does need  data
% with a few days of training we can get the graph
% NN compared to database is lightweighted and time to market
% robustness 

Infomobility covers a fundamental aspect in the smart city ecosystem, and developing corresponding solutions is part of a rapidly evolving software production process. Open Trip Planners are software solutions available on the open-source panorama, allowing public administrations to offer information services to commuters with minimal development effort. In order to enable OTP, data is the first requirement, and these platforms are often based on a common interchange format used for travel agencies, \ie GTFS Transit feed specifications may be unavailable for multiple reasons, \eg technical problems, vendor lock in, different available export formats, data incompatibility, to name a few. We presented in this paper an ML approach to prediction of routes and stops for urban transportation,
%to feed specification 
%({\bf what does it mean here ti feed specification?}), 
%if not available, 
so as to reconstruct the transit graph based on GPS real time tracking data. %These devices are mounted on the fleet and transit bus position, 
%The collected data is processed by an LSTM to produce transit feed specification in the GTFS format. 
%Our approach has been successfully validated on data collected for a single bus, and we plan to test soon the overall process on an entire fleet that will be equipped with GPS tracking devices. 
The obtained accuracy %of our approach 
is quite satisfactory, also considering the initial experiments producing the trips information that we need to refine as part of the future work. More importantly, by measuring the execution time on the dataset, we demonstrate that the framework can process real-time data. %We anticipate that with more representative training data, the proposed model is able to forecast also arrival time. 
In this respect, we suppose that it is ready to be applied to the bus service of the city of L’Aquila. Furthermore, the approach can be easily and successfully replicated to produce transit feed specification for other cities interested in this technology transfer process, contributing also to our experimentation phase.

\IEEEpeerreviewmaketitle



\section*{Acknowledgments} 
	This work has been supported by \emph{(i)} the CROSSMINER Project, EU Horizon 2020 Research and Innovation Programme, grant agreement No. 732223; \emph{(ii)} the research and innovation project n.55/2018 - prot. GSSI nr. 3843/2018, granted by the Municipality of L'Aquila to the GSSI. We would like to thank Giuseppe Bianchi for cooperating in the development of the proposed framework supported by the CUIM project. Finally, we are grateful to the anonymous reviewers for their comments and suggestions, which help us a lot to improve the paper during the revision.	

%Phuong Nguyen gratefully acknowledges the financial supports from the CROSSMINER Project, EU Horizon 2020 Research and Innovation Programme, grant agreement No. 732223.



\bibliographystyle{IEEEtran}
\bibliography{main}

\begin{thebibliography}{10}
\providecommand{\url}[1]{#1}
\csname url@samestyle\endcsname
\providecommand{\newblock}{\relax}
\providecommand{\bibinfo}[2]{#2}
\providecommand{\BIBentrySTDinterwordspacing}{\spaceskip=0pt\relax}
\providecommand{\BIBentryALTinterwordstretchfactor}{4}
\providecommand{\BIBentryALTinterwordspacing}{\spaceskip=\fontdimen2\font plus
\BIBentryALTinterwordstretchfactor\fontdimen3\font minus
  \fontdimen4\font\relax}
\providecommand{\BIBforeignlanguage}[2]{{%
\expandafter\ifx\csname l@#1\endcsname\relax
\typeout{** WARNING: IEEEtran.bst: No hyphenation pattern has been}%
\typeout{** loaded for the language `#1'. Using the pattern for}%
\typeout{** the default language instead.}%
\else
\language=\csname l@#1\endcsname
\fi
#2}}
\providecommand{\BIBdecl}{\relax}
\BIBdecl

\bibitem{qtraspo}
``Special eurobarometer 422a, quality of transport report,'' \url{www.
  data.europa.eu}, 201, p.5.

\bibitem{BOROLE2013775}
\BIBentryALTinterwordspacing
N.~Borole, D.~Rout, N.~Goel, P.~Vedagiri, and T.~V. Mathew, ``Multimodal public
  transit trip planner with real-time transit data,'' \emph{Procedia - Social
  and Behavioral Sciences}, vol. 104, pp. 775 -- 784, 2013, 2nd Conference of
  Transportation Research Group of India (2nd CTRG). [Online]. Available:
  \url{http://www.sciencedirect.com/science/article/pii/S1877042813045631}
\BIBentrySTDinterwordspacing

\bibitem{WATKINS2011839}
\BIBentryALTinterwordspacing
K.~E. Watkins, B.~Ferris, A.~Borning, G.~S. Rutherford, and D.~Layton, ``Where
  is my bus? impact of mobile real-time information on the perceived and actual
  wait time of transit riders,'' \emph{Transportation Research Part A: Policy
  and Practice}, vol.~45, no.~8, pp. 839 -- 848, 2011. [Online]. Available:
  \url{http://www.sciencedirect.com/science/article/pii/S0965856411001030}
\BIBentrySTDinterwordspacing

\bibitem{PORTUGAL2018205}
\BIBentryALTinterwordspacing
I.~Portugal, P.~Alencar, and D.~Cowan, ``The use of machine learning algorithms
  in recommender systems: A systematic review,'' \emph{Expert Systems with
  Applications}, vol.~97, pp. 205 -- 227, 2018. [Online]. Available:
  \url{http://www.sciencedirect.com/science/article/pii/S0957417417308333}
\BIBentrySTDinterwordspacing

\bibitem{Domingos:2012:FUT:2347736.2347755}
P.~Domingos, ``A few useful things to know about machine learning,''
  \emph{Commun. ACM}, vol.~55, no.~10, pp. 78--87, Oct. 2012.

\bibitem{doi:10.1080/21693277.2016.1192517}
T.~Wuest, D.~Weimer, C.~Irgens, and K.-D. Thoben, ``Machine learning in
  manufacturing: advantages, challenges, and applications,'' \emph{Production
  \& Manufacturing Research}, vol.~4, no.~1, pp. 23--45, 2016.

\bibitem{doi:10.1080/03081060.2017.1314502}
\BIBentryALTinterwordspacing
R.~Dalumpines and D.~M. Scott, ``Making mode detection transferable: extracting
  activity and travel episodes from gps data using the multinomial logit model
  and python,'' \emph{Transportation Planning and Technology}, vol.~40, no.~5,
  pp. 523--539, 2017. [Online]. Available:
  \url{https://doi.org/10.1080/03081060.2017.1314502}
\BIBentrySTDinterwordspacing

\bibitem{Ermagun:2017}
A.~Ermagun, Y.~Fan, J.~Wolfson, G.~Adomavicius, and K.~Das, ``Real-time trip
  purpose prediction using online location-based search and discovery
  services,'' Apr. 2017.

\bibitem{Lin2017DeepGM}
Z.~Lin, M.~Yin, S.~Feygin, M.~S. Transportation, J.-F. Paiement, and A.~P. CEE,
  ``Deep generative models of urban mobility,'' 2017.

\bibitem{Zhang0SZ18}
H.~Zhang, H.~Wu, W.~Sun, and B.~Zheng, ``Deeptravel: a neural network based
  travel time estimation model with auxiliary supervision,'' in
  \emph{Proceedings of the 27th Int. Joint Conf. on Artificial Intelligence,
  {IJCAI} July 13-19, 2018, Stockholm, Sweden}, 2018, pp. 3655--3661.

\bibitem{8516374}
J.~{Pang}, J.~{Huang}, Y.~{Du}, H.~{Yu}, Q.~{Huang}, and B.~{Yin}, ``Learning
  to predict bus arrival time from heterogeneous measurements via recurrent
  neural network,'' \emph{IEEE Transactions on Intelligent Transportation
  Systems}, vol.~20, no.~9, pp. 3283--3293, 2019.

\bibitem{Agafonov2019222}
A.~Agafonov and A.~Yumaganov, ``Bus arrival time prediction using recurrent
  neural network with lstm architecture,'' vol.~28, no.~3, pp. 222--230, 2019.

\bibitem{Liu202011917}
H.~Liu, H.~Xu, Y.~Yan, Z.~Cai, T.~Sun, and W.~Li, ``Bus arrival time prediction
  based on lstm and spatial-temporal feature vector,'' vol.~8, pp.
  11\,917--11\,929, 2020.

\bibitem{Zhou2019}
X.~Zhou, P.~Dong, J.~Xing, and P.~Sun, ``Learning dynamic factors to improve
  the accuracy of bus arrival time prediction via a recurrent neural network,''
  vol.~11, 2019.

\bibitem{9013316}
E.~F. d.~S.~{Soares}, H.~{Salehinejad}, C.~A.~V. {Campos}, and S.~{Valaee},
  ``Recurrent neural networks for online travel mode detection,'' in \emph{2019
  IEEE Global Communications Conference (GLOBECOM)}, 2019, pp. 1--6.

\bibitem{YAZDIZADEH201982}
\BIBentryALTinterwordspacing
A.~Yazdizadeh, Z.~Patterson, and B.~Farooq, ``An automated approach from gps
  traces to complete trip information,'' \emph{International Journal of
  Transportation Science and Technology}, vol.~8, no.~1, pp. 82 -- 100, 2019.
  [Online]. Available:
  \url{http://www.sciencedirect.com/science/article/pii/S2046043018300236}
\BIBentrySTDinterwordspacing

\bibitem{8648298}
L.~{Xiao}, Y.~{Zhang}, Z.~{Hu}, and J.~{Dai}, ``Performance benefits of robust
  nonlinear zeroing neural network for finding accurate solution of lyapunov
  equation in presence of various noises,'' \emph{IEEE Transactions on
  Industrial Informatics}, vol.~15, no.~9, pp. 5161--5171, 2019.

\bibitem{XIAO2019124}
\BIBentryALTinterwordspacing
L.~Xiao, Y.~Zhang, J.~Dai, K.~Chen, S.~Yang, W.~Li, B.~Liao, L.~Ding, and
  J.~Li, ``A new noise-tolerant and predefined-time znn model for
  time-dependent matrix inversion,'' \emph{Neural Networks}, vol. 117, pp. 124
  -- 134, 2019. [Online]. Available:
  \url{http://www.sciencedirect.com/science/article/pii/S0893608019301376}
\BIBentrySTDinterwordspacing

\bibitem{Jabamony2020312}
J.~Jabamony and G.~Shanmugavel, ``Iot based bus arrival time prediction using
  artificial neural network (ann) for smart public transport system (spts),''
  vol.~13, no.~1, pp. 312--323, 2020.

\bibitem{Li2019}
H.-M. Li, J.-M. Wu, D.-H. Sun, D.~Chen, and M.~Zhao, ``Bus travel time
  prediction method based on rfid electronic license plate data,'' vol.~32,
  no.~8, pp. 165--173 and 182, 2019.

\bibitem{Petersen2019426}
\BIBentryALTinterwordspacing
N.~C. Petersen, F.~Rodrigues, and F.~C. Pereira, ``Multi-output bus travel time
  prediction with convolutional {LSTM} neural network,'' \emph{Expert Syst.
  Appl.}, vol. 120, pp. 426--435, 2019. [Online]. Available:
  \url{https://doi.org/10.1016/j.eswa.2018.11.028}
\BIBentrySTDinterwordspacing

\bibitem{EY-smartcities}
``Smart city index 2020 report,''
  \url{https://assets.ey.com/content/dam/ey-sites/ey-com/it_it/generic/generic-content/ey_smartcityindex_sostenibilita_marzo2020.pdf},
  accessed: 2020-05-18.

\bibitem{nielsenneural}
\BIBentryALTinterwordspacing
M.~A. Nielsen, ``Neural networks and deep learning,'' 2018. [Online].
  Available: \url{http://neuralnetworksanddeeplearning.com/}
\BIBentrySTDinterwordspacing

\bibitem{Svozil1997}
D.~Svozil, V.~Kvasnicka, and J.~Pospíchal, ``Introduction to multi-layer
  feed-forward neural networks,'' \emph{Chemometrics and Intelligent Laboratory
  Systems}, vol.~39, pp. 43--62, 11 1997.

\bibitem{Bishop:1995:NNP:525960}
C.~M. Bishop, \emph{Neural Networks for Pattern Recognition}.\hskip 1em plus
  0.5em minus 0.4em\relax New York, NY, USA: Oxford University Press, Inc.,
  1995.

\bibitem{7961718}
R.~{Ranjan}, S.~{Sankaranarayanan}, C.~D. {Castillo}, and R.~{Chellappa}, ``An
  all-in-one convolutional neural network for face analysis,'' in \emph{2017
  12th IEEE Int. Conf. on Automatic Face Gesture Recognition (FG 2017)}, May
  2017, pp. 17--24.

\bibitem{DUONG2020105326}
\BIBentryALTinterwordspacing
L.~T. {Duong}, P.~T. {Nguyen}, C.~{Di Sipio}, and D.~{Di Ruscio}, ``{Automated
  fruit recognition using EfficientNet and MixNet},'' \emph{Computers and
  Electronics in Agriculture}, vol. 171, p. 105326, 2020. [Online]. Available:
  \url{http://www.sciencedirect.com/science/article/pii/S0168169919319787}
\BIBentrySTDinterwordspacing

\bibitem{RePEc:eee:intfor:v:14:y:1998:i:1:p:35-62}
G.~Zhang, B.~Eddy~Patuwo, and M.~Y.~Hu, ``Forecasting with artificial neural
  networks:: The state of the art,'' \emph{International Journal of
  Forecasting}, vol.~14, no.~1, pp. 35--62, 1998.

\bibitem{DBLP:conf/aaai/AlemanyBPG19}
\BIBentryALTinterwordspacing
S.~Alemany, J.~Beltran, A.~P{\'{e}}rez, and S.~Ganzfried, ``Predicting
  hurricane trajectories using a recurrent neural network,'' in \emph{The
  Thirty-Third Conference on Artificial Intelligence, {AAAI} 2019, The Ninth
  Symposium on Educational Advances in Artificial Intelligence, {EAAI}
  2019}.\hskip 1em plus 0.5em minus 0.4em\relax {AAAI} Press, 2019, pp.
  468--475. [Online]. Available:
  \url{https://doi.org/10.1609/aaai.v33i01.3301468}
\BIBentrySTDinterwordspacing

\bibitem{10.1162/neco.1997.9.8.1735}
\BIBentryALTinterwordspacing
S.~Hochreiter and J.~Schmidhuber, ``Long short-term memory,'' \emph{Neural
  Comput.}, vol.~9, no.~8, p. 1735–1780, Nov. 1997. [Online]. Available:
  \url{https://doi.org/10.1162/neco.1997.9.8.1735}
\BIBentrySTDinterwordspacing

\bibitem{olah2020}
\BIBentryALTinterwordspacing
C.~Olah, ``{Understanding LSTM Networks},'' May 2020. [Online]. Available:
  \url{https://colah.github.io/posts/2015-08-Understanding-LSTMs/}
\BIBentrySTDinterwordspacing

\bibitem{Shi2019}
X.~Shi, X.~Shao, Z.~Guo, G.~Wu, H.~Zhang, and R.~Shibasaki, ``Pedestrian
  trajectory prediction in extremely crowded scenarios,'' \emph{Sensors},
  vol.~19, p. 1223, 03 2019.

\bibitem{10.1162/neco_a_00990}
\BIBentryALTinterwordspacing
W.~Rawat and Z.~Wang, ``Deep convolutional neural networks for image
  classification: A comprehensive review,'' \emph{Neural Comput.}, vol.~29,
  no.~9, p. 2352–2449, Sep. 2017. [Online]. Available:
  \url{https://doi.org/10.1162/neco_a_00990}
\BIBentrySTDinterwordspacing

\bibitem{Garg:18}
N.~{Garg}, G.~{Ramadurai}, and S.~{Ranu}, ``Mining bus stops from raw gps data
  of bus trajectories,'' in \emph{2018 10th International Conference on
  Communication Systems Networks (COMSNETS)}, Jan 2018.

\bibitem{gmd-7-1247-2014}
\BIBentryALTinterwordspacing
T.~Chai and R.~R. Draxler, ``Root mean square error (rmse) or mean absolute
  error (mae)? – arguments against avoiding rmse in the literature,''
  \emph{Geoscientific Model Development}, vol.~7, no.~3, pp. 1247--1250, 2014.
  [Online]. Available: \url{https://www.geosci-model-dev.net/7/1247/2014/}
\BIBentrySTDinterwordspacing

\bibitem{tiwari2010appraisal}
R.~Tiwari, M.~Arora, and A.~Kumar, ``An appraisal of gps related errors,''
  \emph{Geospatial World}, 2010.

\end{thebibliography}

\end{document}